\documentclass[conference]{IEEEtran}

\usepackage{tikz}
\usepackage{algorithm}
\usepackage{algpseudocode}
\usepackage{xcolor}
\usepackage[skip=0pt]{subcaption}
\usepackage{amsmath,amsfonts}
\usepackage{hyperref}
\usepackage{graphicx}
\usepackage{multirow}
\usepackage{amsthm}
\usepackage{balance}
\usepackage{float}
\usepackage{booktabs}
\usepackage{makecell}
\usepackage{colortbl}
\usepackage[T1]{fontenc}
\usepackage{pifont}

\usepackage{enumitem}
\setlist{leftmargin=0.6cm}

\newtheorem{theorem}{Theorem}
\newtheorem{proposition}{Proposition}
\newtheorem{lemma}{Lemma}
\newtheorem{claim}{Claim}
\newtheorem{example}{Example}
\newtheorem{remark}{Remark}

\newtheorem{definition}{Definition}
\newtheorem{corollary}{Corollary}

\newcommand{\fmgftoc}{-0.2cm}
\newcommand{\fmgctom}{-0.3cm}

\newcommand{\notbf}{\noindent\textbf}

\newcommand{\equsize}{\small}
\newcommand{\equienvs}{\begin{normalsize}}
\newcommand{\equienve}{\end{normalsize}}
\newcommand{\parastart}{\noindent\textbf}

\newcommand{\revisestart}{\begin{color}{black}}
\newcommand{\reviseend}{\end{color}}

\newcommand{\RevisionChangeStart}{\begin{color}{black}}
\newcommand{\RevisionChangeEnd}{\end{color}}

\ifCLASSINFOpdf
\else
\fi
\begin{document}


\pagestyle{plain} 
%
\title{Revisiting Differentially Private Hyper-parameter Tuning}

\author{

\IEEEauthorblockN{Zihang Xiang}
\IEEEauthorblockA{KAUST\\zihang.xiang@kaust.edu.sa}

\and

\IEEEauthorblockN{Tianhao Wang}
\IEEEauthorblockA{University of Virginia\\tianhao@virginia.edu}

\and

\IEEEauthorblockN{Cheng-Long Wang}
\IEEEauthorblockA{KAUST\\chenglong.wang@kaust.edu.sa}

\and

\IEEEauthorblockN{Di Wang}
\IEEEauthorblockA{KAUST\\di.wang@kaust.edu.sa}

}

\maketitle


\begin{abstract}

We investigate the application of differential privacy in hyper-parameter tuning, a process involving selecting the best run from several candidates. Unlike many private learning algorithms, including the prevalent DP-SGD, the privacy implications of selecting the best are often overlooked. While recent works propose a generic \textit{private selection} solution for the tuning process, an open question persists: is such privacy upper bound tight?

This paper provides both empirical and theoretical examinations of this question. Initially, we provide studies affirming the current privacy analysis for private selection is indeed tight in general. However, when we specifically study the hyper-parameter tuning problem in a white-box setting, such tightness no longer holds. This is first demonstrated by applying privacy audit on the tuning process. Our findings underscore a substantial gap between the current theoretical privacy bound and the empirical privacy leakage derived even under strong audit setups.

This gap motivates our subsequent theoretical investigations, which provide improved privacy upper bound for private hyper-parameter tuning due to its distinct properties. Our improved bound leads to better utility. Our analysis also demonstrates broader applicability compared to prior analyses, which are limited to specific parameter configurations. Overall, we contribute to a better understanding of how privacy degrades due to \textit{selection} \footnote{To appear in Network and Distributed System Security (NDSS) Symposium 2026}.

\end{abstract}


%
\IEEEpeerreviewmaketitle

\section{Introduction}
Differential Privacy (DP) \cite{DBLP:conf/tcc/DworkMNS06} stands as the prevailing standard for ensuring privacy in contemporary machine learning. A ubiquitous technique employed to ensure DP across a diverse array of machine learning tasks is differentially private stochastic gradient descent (DP-SGD, {\it a.k.a.}, noisy-SGD) \cite{bassily2014private, song2013stochastic, abadi2016deep}.

In addition to a single (private) training process, machine learning systems always involve a hyper-parameter tuning process that entails running a (private) \textit{base} algorithm (e.g., DP-SGD) multiple times with different configurations and selecting the best run. Regrettably, unlike the well-studied DP-SGD, the reasoning for the privacy cost of such tuning operations is inadequately studied and often totally ignored. 

Naively, one can bound the privacy loss for the tuning operation by the composition theorem. If we run the private base algorithm $k$ times with different hyper-parameters, the total privacy cost deteriorates at most linearly with $k$ (or $\mathcal{O}(\sqrt{k})$ if the base algorithm is approximate DP and we use advanced composition theorem \cite{dwork2014algorithmic}, which is nearly optimal~\cite{kairouz2015composition}). However, these bounds are still far from satisfactory as $k$ is usually large in practice. 
Perhaps due to this limitation, it still remains common to exhaustively tune a private algorithm to achieve strong performance but only consider the privacy cost for a single run \cite{de2022unlocking,xiao2023theory,xiang2023practical,shen2023differentially}.

For another approach, tuning hyper-parameter privately can be framed as a \textit{private selection} problem, for which several well-explored mechanisms, such as the sparse vector technique \cite{dwork2009complexity} and the exponential mechanism \cite{mcsherry2007mechanism}, may potentially be utilized. However, these mechanisms assume that the score function (defining the ``best'' to be selected) has low sensitivity (for DP analysis), which is a condition not always met. 

Thanks to Liu and Talwar \cite{liu2019private}, hyper-parameter tuning now enjoys significantly better privacy bound than naively applying composition theorem. To briefly describe their findings, if we run a private base algorithm a \textit{random} number of times (possibly with different hyper-parameters) and only output the best single run, the privacy cost only deteriorates by a constant multiplicative factor \cite{liu2019private}. For example, if the base algorithm is $(\varepsilon,0)$-DP, then the whole tuning process is $(3\varepsilon,0)$-DP if the running number follows a geometric distribution \cite{liu2019private}. This is much better than the $(k\varepsilon,0)$-DP bound under \textit{fixed} $k$ times of running. 

Later, Papernot and Steinke \cite{papernot2021hyperparameter} operate within R\'enyi DP (RDP) framework~\cite{mironov2017renyi} and mandate the randomization of the number of running times, presenting additional results for varying degrees of randomness. A noteworthy aspect of both methodologies \cite{liu2019private, papernot2021hyperparameter} lies in their treatment of the base algorithm as a \textit{black box}; thus, such a generic approach applies to a broader spectrum of private selection problems, provided the base algorithm is differentially private on its own.

\parastart{{Motivations}.} 
\RevisionChangeStart
Hyper-parameters can be tuned with formal privacy guarantees. But it remains unclear whether the existing privacy analyses~\cite{liu2019private,papernot2021hyperparameter} are tight.

Some results suggest the cost of tuning is high. For example, \cite{liu2019private} shows the privacy budget can increase by a factor of three in certain setups. Still, this seems counterintuitive. It seems plausible that \textit{only} revealing the best \textit{single} run should not consume that much privacy budget. This leads to our core question:
Does hyperparameter tuning actually consume significantly more privacy than the base algorithm?

If the answer is positive, then further significant improvements in the analysis is impossible. If negative, it is still valuable to pursue tighter analysis, as tighter bounds would allow more hyperparameter trials under the same budget, directly improving utility in private model selection.
\RevisionChangeEnd

\parastart{{This work}.} We answer the posed question with both positive and negative answers. In the affirmative, our constructed example demonstrates that the current generic privacy bound provided in \cite{papernot2021hyperparameter} for private selection is indeed tight. Still, the result only holds in the worst case. Conversely, in the negative, we uncover a more favorable privacy bound given the base algorithm is specific DP-SGD. We aim to understand how \underline{\textit{selection}} leaks privacy, in contrast to the well-established understanding of privacy deterioration due to \textit{composition}. Our contributions are as follows.

\notbf{1) Validating tightness of generic privacy bound for private selection (Section \ref{sec:validate_tight}).} We first provide a private selection instance where we observe only a negligible gap between the true privacy cost and the cost predicted by the current privacy bound \cite{papernot2021hyperparameter}. Such results meaningfully show that significant improvement in privacy upper bound \textit{is impossible in general}.
However, when we study the private hyper-parameter tuning problem, where the base algorithm is DP-SGD, we enjoy better upper bounds. This finding is related to our other two contributions.

\notbf{2) Empirical investigation on how hyper-parameter tuning (selection) leaks privacy (Section \ref{sec:exp}).} We first take empirical approaches to investigate how much privacy is leaked when performing hyper-parameter tuning. This is done via the privacy audit technique~\cite{nasr2021adversary, kairouz2015composition, jagielski2020auditing}, an interactive protocol used to empirically measure the privacy of some mechanisms. 

In contrast, unlike all previous privacy auditing work, which focuses on the privacy of the base algorithm (e.g., DP-SGD), auditing the tuning procedure is a fresh problem that requires new formulation and insight. Specifically, the score function used to select the ``best'' is the new factor that must be settled.

We formulate various privacy threat models tailored for hyper-parameter tuning, where the weakest one corresponds to the most practical scenario and the strongest one corresponds to the worst case. Results under the weakest provide evidence that the tuning process hardly incurs additional privacy costs beyond the base algorithm. 
Moreover, even the empirical privacy bound (lower bound) derived from the strongest adversary still exhibits a substantial gap from the generic privacy bound (upper bound) proposed by \cite{papernot2021hyperparameter}. 
    \textit{In contrast, previously, the gap (between privacy lower bound and upper bound) is essentially closed in the audit on DP-SGD's privacy \cite{nasr2021adversary}.}
Why are different audit results seen in auditing hyper-parameter tuning? This motivates our subsequent theoretical investigations.

\notbf{3) Improved theoretical privacy results (Sections \ref{sec:imp_fdp}, \ref{sec:bound_disc} and \ref{sec:further_eval}).} Our subsequent study shows that tuning DP-SGD does enjoy a better privacy result. The pivotal aspect driving this improvement lies in representing the privacy of the base algorithm with finer resolution, and DP-SGD does have a distinctive characterization. This is done within the $f$-DP framework \cite{dong2019gaussian}, deviating from the well-known $(\varepsilon,\delta)$-DP \cite{DBLP:conf/tcc/DworkMNS06} or RDP \cite{mironov2017renyi}. 

Our improved result directly benefits differentially private hyper-parameter tuning: 

    \textit{it allows us to test substantially more (in expectation) hyper-parameters without increasing privacy budget, which translates to improved utility}.

Our results are generalizable, contrasting to previous work \cite{liu2019private,papernot2021hyperparameter}, which remains unknown how to adapt to general parameter setups. 

Subsequent to our improved results is a further empirical evaluation: comparing our improved theoretical privacy result with the empirical privacy lower bound derived under an idealized audit setup. Interestingly, there is still a gap in between. This finding is examined in detail, revealing that the score function, a new factor in auditing hyper-parameter tuning, is a key determinant influencing audit performance. Consequently, this also prompts an exciting and essential open problem in the future: how to close such a gap.

\section{Background}
\subsection{Differential Privacy (DP)} 
\begin{definition}[Differential Privacy \cite{DBLP:conf/tcc/DworkMNS06}]\label{def:dp}
Given a data universe $\mathcal{X}$, two datasets $X, X'\subseteq \mathcal{X}$ are adjacent if they differ by one data example. A randomized algorithm $\mathcal{M}$ satisfies $(\varepsilon,\delta)$-differential privacy, or $(\varepsilon,\delta)$-DP, if for all adjacent datasets $X,\, X'$ and for all events $S$ in the output space of $\mathcal{M}$, we have $\operatorname{Pr}(\mathcal{M}(X)\in S)\leq e^{\varepsilon} \operatorname{Pr}(\mathcal{M}(X')\in S)+\delta$.
\end{definition}

We introduce R\'enyi DP (RDP), a DP relaxation shown in the following, often serves as a tight analytical tool to assess the privacy cost under composition.

\begin{definition}[R\'enyi DP \cite{DBLP:conf/csfw/Mironov17}]\label{def:RDP} The R\'enyi divergence is defined as $\mathcal{D}_{\alpha}(M||N) = \frac{1}{\alpha - 1} \ln \mathbb{E}_{x\sim N}\left[\frac{M(x)}{N(x)}\right]^\alpha$
with $\alpha>1$. 
A randomized mechanism $\mathcal{M}: \mathcal{X}\rightarrow \mathcal{Y} $ is said to be $(\alpha, \gamma)$-R\'enyi DP, or $(\alpha, \gamma)$-RDP, if $\mathcal{D}_{\alpha}(\mathcal{M}(X)||\mathcal{M}(X')) \leq \gamma$
holds for any adjacent dataset $X, X'$.

\end{definition}

\notbf{Differentially private stochastic gradient descent (DP-SGD) \cite{bassily2014private, song2013stochastic, abadi2016deep}.} 
\RevisionChangeStart
We use a machine learning model $f_w$, typically a neural network with trainable parameters $w$. In our classification setting, $f_w$ maps inputs (e.g., images) to labels. Parameters are updated using Stochastic Gradient Descent (SGD) \cite{lecun1998gradient}, where hyper-parameters like learning rate must be tuned for good performance.

DP-SGD is the private version of SGD. It follows three steps:
1) compute per-sample gradients;
2) clip each to have bounded $\ell_2$ norm;
3) add Gaussian noise.

The private gradient $p_i$ is then used to update $w$ as:
\equienvs
\begin{equation}\label{equ:dpsgd}
\begin{aligned}
    &p_i = \sum_{(x, y) \in \mathbf{B}} \operatorname{CLP}_C\left(\nabla_w \ell\left(w_{i-1}; x, y\right)\right)+R_i\\
    &w_{i} \leftarrow w_{i-1}-\mathbf{lr} \cdot p_i
\end{aligned}
\end{equation}
\equienve
 
Here, $\mathbf{B}$ is the sampled batch (with ratio $\tau$), $\mathbf{lr}$ is the learning rate, and $\ell$ is the loss (e.g., cross-entropy). Clipping is defined as $\operatorname{CLP}_C(u) = u \cdot \min(1, \frac{C}{|u|_2})$, where $C$ is a clipping threshold. Noise $R_i$ is sampled from $\mathcal{N}(0, C^2\sigma^2 \mathbb{I}^d)$, where $\sigma$ is the noise multiplier and $d$ is the number of parameters. Removing clipping and noise recovers (mini-batch) SGD. Variants like DP-Adam \cite{tang2023dp} follow the same privacy analysis by the post-processing property of DP.
\RevisionChangeEnd

\subsection{Privacy Audit}\label{sec:audit_intro}
\notbf{Hypothesis testing interpretation of DP.}
For a randomized mechanism $\mathcal{M}$, let $X,X'$ be adjacent datasets, let $y\in \mathcal{Y}$ be the output of $\mathcal{M}$ taking input $X$ or $X'$, we form the \textit{null}
and \textit{alternative} hypotheses:
\equienvs
\begin{equation}\label{equ:basic_hypo}
\begin{aligned}
\mathbf{H_0}:\text{$X$ was the input},\text{\quad}\mathbf{H_1}:\text{$X'$ was the input}. 
\end{aligned}
\end{equation}
\equienve
For any decision rule $\mathcal{R}:\mathcal{Y}\rightarrow \{0,1\}$ in such a hypothesis testing problem, it has two notable types of errors: 1) type I error or false positive rate $\mathrm{FP}=\operatorname{Pr}(\mathcal{R}(y)=1|\mathbf{H_0})$, i.e., the probability of rejecting $\mathbf{H_0}$ while $\mathbf{H_0}$ is true; 2) type II error or false negative rate $\mathrm{FN}=\operatorname{Pr}(\mathcal{R}(y)=0|\mathbf{H_1})$, i.e., the probability of rejecting $\mathbf{H_1}$ while $\mathbf{H_1}$ is true. DP can be characterized by such two error rates as follows.
\begin{theorem}[DP as Hypothesis Testing \cite{kairouz2015composition}]\label{thm:hp_dp} For any $\varepsilon>0$ and $\delta\in[0,1]$, a mechanism $\mathcal{M}$ is $(\varepsilon,\delta)$-DP if and only if
\equienvs
\begin{equation}\label{equ:privacy_region}
\begin{aligned}
    & \mathrm{FP}+\mathrm{e}^\varepsilon \mathrm{FN}\geq 1-\delta, \text{\quad} \mathrm{FN}+\mathrm{e}^\varepsilon \mathrm{FP}\geq 1-\delta
\end{aligned}
\end{equation}
\equienve
both hold for any adjacent dataset $X,X'$ and any decision rule $\mathcal{R}$ in a hypothesis testing problem as defined in Equation \eqref{equ:basic_hypo}.
\end{theorem}
Theorem \ref{thm:hp_dp} has the following implications. With $\delta$ fixed at some value, under the threat model that an adversary can only operate at some $\mathrm{FP}$ and $\mathrm{FN}$ under some decision rule $\mathcal{R}$ for a specific adjacent dataset pair $X,X'$, a \textit{lower bound}
\equienvs
\begin{equation}\label{equ:lower_bound}
\begin{aligned}
    \varepsilon_L^{(X,X',\mathcal{R})}=\max
    \{\log \frac{1-\delta-\mathrm{FP}}{\mathrm{FN}}, \log \frac{1-\delta-\mathrm{FN}}{\mathrm{FP}}, 0\}
\end{aligned}
\end{equation}
\equienve
can be computed, meaning that the algorithm cannot be more private than that, i.e., the true privacy parameter $\varepsilon_T\geq\varepsilon_L^{(X,X',\mathcal{R})}$,
just as entailed by Theorem \ref{thm:hp_dp}. Finding $\varepsilon_T$ requires taking the maximum of lower bound value over all pairs of $X,X'$ and $\mathcal{R}$, which is clearly intractable in general. In practice, people are satisfied by reporting an \textit{upper bound} $\varepsilon_U\geq\varepsilon_T$, which is obtained by analytical approaches (privacy accounting) \cite{abadi2016deep, mironov2017renyi,rdp_subgaussian_mir}.

\begin{algorithm}[!ht]
\caption{
Game-based Privacy Audit $\mathcal{G}$
}\label{alg:audit}

\begin{algorithmic}[1]
\equsize
\renewcommand{\algorithmicrequire}{\textbf{Input:}}
\renewcommand{\algorithmicensure}{\textbf{Output:}}

\Require {DP protocol $\mathcal{P}$, adjacent pair $X, X'$}
\State $b_{\text{truth}} \gets \{0,1\}$ \Comment{\textit{Trainer} flips a fair coin}
\State $\hat{X}\gets X$ if $b_{\text{truth}}=0$, $\hat{X}\gets X'$ otherwise
\State Run $\mathcal{P}(\hat{X})$ \Comment{\textit{Trainer} runs the private protocol}\label{alg:audit_obs}
\State $b_{\text{guess}} \gets \{0,1\}$ \label{alg:audit_make_assertion} \Comment{\textit{Adversary makes a guess based on $\mathcal{P}(\hat{X})$}}

\Ensure $(b_{\text{truth}}, b_{\text{guess}})$

\end{algorithmic}
\end{algorithm}

\notbf{Privacy audit}. Privacy audit aims to find a lower bound of the privacy cost for a private protocol $\mathcal{P}$ based on the hypothesis testing interpretation of DP as shown above. This is usually done via simulating the interactive game-based protocol described in  Algorithm \ref{alg:audit}.
Such a simulation is typically repeated many times, resulting in many pairs of $(b_{\text{truth}},b_{\text{guess}})$. Then, the $\mathrm{FP}$ and $\mathrm{FN}$  for adversary's guessing are computed by Clopper-Pearson method \cite{clopper1934use} with a confidence specification.
If the adversary can make very accurate guesses and derive a lower bound higher than some claimed privacy parameter, it suggests $\mathcal{P}$ is not private as claimed.

\RevisionChangeStart
Audit only gives a lower bound 
$\varepsilon_L^{(X,X',\mathcal{R})}$ of the true privacy bound $\varepsilon_T$, meaning that the algorithm is at least not $(\varepsilon_L^{(X,X',\mathcal{R})},\delta)$. The lower bound due to privacy audit is different from the upper bound given by theory. The limitation of privacy audit is that the result it gives should not be used as a formal privacy guarantee.
\RevisionChangeEnd

\notbf{Related work on privacy audit.} 
In privacy-preserving machine learning, privacy audit mainly serves a different goal from that of 
certain earlier studies \cite{wang2020checkdp, ding2018detecting,bichsel2021dp, bichsel2018dp} on detecting privacy violation in general query-answering applications. Previous work on privacy audit in machine learning mainly targets auditing the DP-SGD protocol to assess its theoretical versus practical privacy \cite{nasr2021adversary, kairouz2015composition, jagielski2020auditing}. Additional studies \cite{steinke2023privacy, nasr2023tight, lu2022general, zanella2023bayesian} concentrate on enhancing the strength of audits on DP-SGD (yielding stronger/larger-value lower bound) or improving the efficiency (incurring fewer simulation overheads). Drawing a parallel to the action of guessing whether a data point was included or not, privacy audit may also be linked to \textit{membership inference attack} (MIA) \cite{ nasr2019comprehensive,shokri2017membership}. Still, privacy audit aims to give a privacy lower bound. There are also recent works on auditing prediction \cite{chadha2024auditing} and synthetic data generation \cite{annamalai2024you}, which differ from our auditing experiments.

\subsection{Private Hyper-parameter Tuning}
\notbf{Problem Formulation}. We formulate the private hyper-parameter tuning problem aligning with \cite{liu2019private, papernot2021hyperparameter}. Let $\Omega=\{\mathcal{M}_1, \mathcal{M}_2,\cdots, \mathcal{M}_m\}$ be a collection of DP-SGD algorithms ($m$ is chosen freely). These correspond to $m$ possible hyper-parameter configurations. We have $\mathcal{M}_i:\mathcal{X}\rightarrow \mathcal{Y}$ for $i\in[m]$, and all of these algorithms satisfy the same privacy parameter, e.g., they satisfy the same $(\varepsilon,\delta)$-DP guarantee. 

Note that it requires that each $\mathcal{M}_i$ to be differentially private on its own \cite{liu2019private, papernot2021hyperparameter}, meaning that indistinguishability exists between distribution $\mathcal{M}_i(X)$ and $\mathcal{M}_i(X')$, for any $X,X'$ being adjacent and for any $\mathcal{M}_i$'s hyper-parameter.


Finally, it is to return an algorithm element (including its execution) of $\Omega$ such that the output of such algorithm has (approximately) the maximum score as specified by some score function $g:\mathcal{Y}\rightarrow \mathbb{R}$. 
The score function $g$ usually serves a utility purpose (e.g., $g$ could evaluate the validation loss on a held-out dataset). The selection must be performed in a differentially private manner. The general \textit{private selection} problem corresponds to the cases where $\Omega$ contains arbitrary differentially private algorithms.

\notbf{Related work on private hyper-parameter tuning.} Well-known algorithms like the sparse vector technique \cite{dwork2009complexity} and exponential mechanism \cite{mcsherry2007mechanism} may potentially be leveraged to the tuning problem; however, they assume a low sensitivity in the metric defining the ``best'', a condition not always applicable. Some earlier work \cite{chaudhuri2013stability} also suffers from the same issue. Papernot et al. \cite{papernot2021hyperparameter} and Liu and Talwar \cite{liu2019private} have provided generic private selection approaches circumventing such challenges. Mohapatra et al. \cite{mohapatra2022role} study privacy issues in \textit{adaptive} hyper-parameter tuning under DP, which is different from the \textit{non-adaptive} tuning problem considered in this work. There is related work \cite{koskela2023practical} that integrates the solution from \cite{papernot2021hyperparameter} to some larger algorithm; therefore, what we understand about the generic approach in this study naturally propagates to \cite{koskela2023practical}.

\notbf{Focus of this paper.} Towards understanding how selection leaks privacy, our first focus is to formulate specific privacy audit to understand how privacy deteriorates due to \textit{selection}, diverging from all previous privacy audit work on understanding privacy deteriorating due to \textit{composition}.
Also motivated by our empirical findings, we further improve privacy upper bound specifically for a white-box application: the hyper-parameter tuning problem, pre-conditioned on the base algorithm, is DP-SGD. \RevisionChangeStart
There is another notable work on private hyper-parameter tuning \cite{ding2022revisiting} by proposing a different algorithm with different assumptions than \cite{papernot2021hyperparameter} and \cite{liu2019private} (in \cite{papernot2021hyperparameter} and \cite{liu2019private}, they only require the base algorithm to be DP; in \cite{ding2022revisiting}, they must partition the training dataset to be disjoint). Since the starting point of this work is \cite{papernot2021hyperparameter} and \cite{liu2019private}, we will only focus on the same line of \cite{papernot2021hyperparameter} and \cite{liu2019private}.
\RevisionChangeEnd

\section{Current Private Selection Protocol}\label{sec:pri_sel}
\subsection{Current Algorithm}\label{sec:private_tune}
\RevisionChangeStart

We start with the state-of-the-art algorithm for private selection \cite{liu2019private, papernot2021hyperparameter}, shown in Algorithm~\ref{alg:ps}. This generic method applies when the base algorithm is already differentially private. When each $\mathcal{M}_i \in \Omega$ is a DP-SGD instance, the task becomes private hyperparameter tuning. If the base algorithm $\mathcal{M}$ is $(\varepsilon, 0)$-DP and $\xi$ is geometric, then Algorithm~\ref{alg:ps} is $(3\varepsilon, 0)$-DP \cite{liu2019private}. \cite{papernot2021hyperparameter} improves this for pure DP by using a Truncated Negative Binomial (TNB) distribution for $\xi$ under specific parameters (see Appendix~\ref{app:TNBD}). The improvement uses RDP-based analysis.
\RevisionChangeEnd

\begin{algorithm}[!ht]
\caption{Private Selection Protocol $\mathcal{H}$ \cite{papernot2021hyperparameter,liu2019private}}\label{alg:ps}
\begin{algorithmic}[1]
\equsize
\renewcommand{\algorithmicrequire}{\textbf{Input:}}
\renewcommand{\algorithmicensure}{\textbf{Output:}}

\Require {Dataset $X$; algorithms $\Omega$; distribution $\xi$; score function $g$}

\State Draw a sample: $k \leftarrow \xi$
\State  $Y\gets \text{\textbf{Null}}$, $S \gets -\infty$
\For{$i=1,2,\cdots,k$}
    \State Uniformly randomly fetch one element $\mathcal{M}_i$ from $\Omega$\label{alg:ps_instantiation}
    \State  $y_i\gets\mathcal{M}_i(X)$    \Comment{Run $\mathcal{M}_i$ on dataset $X$}
    \State \textbf{If $g(y_i) > S$}: $Y\gets y_i$, $S\gets g(y_i)$ \Comment{Selecting the ``best''}

\EndFor

\Ensure $Y$ 
\end{algorithmic}
\end{algorithm}

\subsection{Our General Tightness Proof}\label{sec:validate_tight}
We show the current privacy upper bound due to \cite{papernot2021hyperparameter} is tight in a general sense.

\begin{example}[Our Construction for Pure DP] \label{example:pure_dp_tight} Let $\mathcal{M}$ have a finite output space $\mathcal{Y}=\{\mathrm{A}, \mathrm{B}, \mathrm{C}\}$.  $\mathcal{M}$ only cares about the number of data samples in its input. If the number is even, its output follows the distribution shown as the left-hand side of Equation \eqref{equ:base_pure_dp}; otherwise, its output distribution is the right-hand side.
\equienvs
\begin{equation}\label{equ:base_pure_dp}
\begin{aligned}
\operatorname{Pr}_\mathcal{M}\left\{\begin{array}{ll}
        \operatorname{Pr}_{\mathrm{A}} =1-b\mathrm{e}^\varepsilon-db\\
        \operatorname{Pr}_{\mathrm{B}} = b\mathrm{e}^\varepsilon \\
        \operatorname{Pr}_{\mathrm{C}} = db\\
\end{array}\right.
\operatorname{Pr}_\mathcal{M'}\left\{\begin{array}{ll}
        \operatorname{Pr}_{\mathrm{A}'} =1-b-db\mathrm{e}^\varepsilon\\
        \operatorname{Pr}_{\mathrm{B}'} =b\\
        \operatorname{Pr}_{\mathrm{C}'} =db\mathrm{e}^\varepsilon \\
\end{array}\right.
\end{aligned}
\end{equation}
\equienve
where $\operatorname{Pr}_{\mathrm{A}}$ denotes the probability of event $A$ occurs conditioned on even (similarly we also have  $\operatorname{Pr}_{\mathrm{A}'}$ with respect to odd). With $b = 10^{-3}, d=10^2, \varepsilon=1$, we can see $\mathcal{M}$ is clearly $(1,0)$-DP for any pair of adjacent (w.r.t. addition/removal) dataset.

Let each element $\mathcal{M}_i$ fetched from $\Omega$ in line \ref{alg:ps_instantiation} of Algorithm \ref{alg:ps} has the same output distribution as Equation \eqref{equ:base_pure_dp}. Also let a score function $g$ give $g(\mathrm{C})>g(\mathrm{B})>g(\mathrm{A})$. Let $\xi$ be the TNB distribution with parameter $\eta=1, \nu=10^{-3}$ (geometric distribution). The probability for each event that Algorithm \ref{alg:ps} outputs is computed by the following.

\begin{claim}\label{thm:prob_compute}
    Let $y$ be some event in $\mathcal{Y}$, the probability of $y$ occurs as the output of the tuning process $\mathcal{H}$ (Algorithm \ref{alg:ps}) is
    \equienvs
    \begin{equation}\label{equ:y_prob}
    \begin{aligned}
        \operatorname{Pr}(y)=\sum_{k\sim\xi}\operatorname{Pr}(k)\left(\operatorname{Pr}(E_{\leq y})^k -\operatorname{Pr}(E_{<y})^k\right),
    \end{aligned}
    \end{equation}
    \equienve
    where $E_{\leq y}=\{x:g(x)\leq g(Y)\}$ and $E_{< y}=\{x:g(x)< g(Y)\}$. See proof in Appendix \ref{app:proof_of_prob_compute}.
\end{claim}
Let $\operatorname{Pr}_\mathcal{H},\operatorname{Pr}_\mathcal{H'}$ denote the probabilities for each event conditioned on $\mathcal{H}$ operates on adjacent dataset pair. For $\operatorname{Pr}_\mathcal{H}$ we have 
\equienvs
\begin{equation}\nonumber
\begin{aligned}
\operatorname{Pr}_\mathcal{H}\left\{\begin{array}{ll}
\operatorname{Pr}_{\mathrm{A}|\mathcal{H}}=\sum_{k\sim\xi}\operatorname{Pr}(k)\operatorname{Pr}_{\mathrm{A}}^k \\
\operatorname{Pr}_{\mathrm{B}|\mathcal{H}}=\sum_{k\sim\xi}\operatorname{Pr}(k)((\operatorname{Pr}_{\mathrm{A}}+\operatorname{Pr}_{\mathrm{B}})^k -\operatorname{Pr}_{\mathrm{A}}^k) \\
\operatorname{Pr}_{\mathrm{C}|\mathcal{H}}=\sum_{k\sim\xi}\operatorname{Pr}(k)(1-(\operatorname{Pr}_{\mathrm{A}}+\operatorname{Pr}_{\mathrm{B}})^k)\\
\end{array}\right.
\end{aligned}
\end{equation}
\equienve
where $\operatorname{Pr}_{\mathrm{A}|\mathcal{H}}$ denotes the probability of event $A$ occurs as the output of $\mathcal{H}$ conditioned on the input dataset contains even number of data points.
$\operatorname{Pr}_\mathcal{H'}$ can be computed similarly. Numerically, this gives the probabilities shown below
\equienvs
\begin{equation}\label{equ:hyper_pure_dp}
\begin{aligned}
\operatorname{Pr}_\mathcal{H}\left\{\begin{array}{ll}
        \operatorname{Pr}_{\mathrm{A}|\mathcal{H}}=8.66\times 10^{-3}\\
        \operatorname{Pr}_{\mathrm{B}|\mathcal{H}}=2.60\times 10^{-4}\\
        \operatorname{Pr}_{\mathrm{C}|\mathcal{H}}=9.91\times 10^{-1}\\
\end{array}\right.
\operatorname{Pr}_\mathcal{H'}\left\{\begin{array}{ll}
        \operatorname{Pr}_{\mathrm{A}|\mathcal{H}'}=2.66\times 10^{-3}\\
        \operatorname{Pr}_{\mathrm{B}|\mathcal{H}'}=1.34\times 10^{-5}\\
        \operatorname{Pr}_{\mathrm{C}|\mathcal{H}'}=9.97\times 10^{-1}\\
\end{array}\right.
\end{aligned}
\end{equation}
\equienve
and it can be checked to satisfy $(2.96,0)$-DP. The theoretical bound claims Algorithm \ref{alg:ps} is $(3,0)$-DP, i.e., it is tight up to a negligible gap.
\end{example}

\RevisionChangeStart
Our example shows non-asymptotic tightness—an exact bound up to negligible error. This is more convincing than the example in \cite[Appendix D.3]{papernot2021hyperparameter}, which relies on assumptions and first-order approximations. For approximate DP ($\delta > 0$), tightness also holds and is shown trivially in Appendix~\ref{app:TNBD}.

This raises a new question: Does this worst-case tightness still apply when tuning hyper-parameters using multiple DP-SGD runs? We explore this in the following sections, focusing on the case where each $\mathcal{M}_i \in \Omega$ is a DP-SGD instance with the same privacy guarantee. To begin, we conduct privacy audit experiments to examine how tight the previous bounds are in practice for DP-SGD in the next section.
\RevisionChangeEnd

\begin{table}[!ht] 
\small
\centering
\resizebox{0.92\columnwidth}{!}{
\begin{tabular}{cl}
\toprule
\toprule

Notation & Meaning\\
\midrule
$\mathcal{G}$ & The distinguishing game, Algorithm \ref{alg:audit}\\
$\mathcal{P}$ & A general protocol to be audited in $\mathcal{G}$\\
$\mathcal{H}$ & The private tuning protocol, Algorithm \ref{alg:ps}\\
$\mathcal{M}$ & The base algorithm (DP-SGD) of $\mathcal{H}$ \\
$\mathbb{F}, \mathbb{M}, \mathbb{C}, \mathbb{S}$ & Datasets used, shown in Section \ref{sec:presentation_eval} \\
$N$ & Number of iterations inside $\mathcal{M}$ \\
$C$ & Clipping threshold in Equation \eqref{equ:dpsgd} \\
\midrule

$w_{i}$ & Model at $i$-th iteration in Equation \eqref{equ:dpsgd}\\
$\ell$ & The loss function in Equation \eqref{equ:dpsgd}\\
$\xi$ & Running number distribution of $\mathcal{H}$\\
$g$ & Score function evaluating $\mathcal{M}$'s output\\
$\mathbf{z}$ & Differing data point, constructed by adversary\\
$p_i^\mathbf{z}$ & $\mathbf{z}$'s gradient at $i$-th iteration in Equation \eqref{equ:dpsgd}\\
$p_i$ & Private gradient in Equation \eqref{equ:dpsgd}\\
$\mathbf{Z}_D$ & Hypothetical $\mathbf{z}$ leading to $\textit{\underline{D}irac}$ gradient\\

\midrule

$\lambda_a, \lambda_b$ & Two proxies constructed by the adversary\\
$\sigma$ & Noise s.t.d. for $R_i$ in Equation \eqref{equ:dpsgd}\\
$\varepsilon_B$ & Base algorithm $\mathcal{M}$'s privacy budget\\
$\varepsilon_L$ & Lower bound for $\mathcal{H}$ by audit\\
$\varepsilon_U$ & Generic upper bound for $\mathcal{H}$, by \cite{papernot2021hyperparameter} \\
\bottomrule
\bottomrule
\end{tabular}

}

\caption{
Notations used in our empirical study. 
}
\label{tab:notations}
\end{table}

\section{Empirical Investigation}\label{sec:exp}
In this section, we aim to find how much privacy is leaked due to the tuning procedure $\mathcal{H}$ when the base algorithm is specifically the DP-SGD protocol. Notations used are summarised in Table \ref{tab:notations}.

\subsection{High-level Procedure}\label{sec:game_formulation}

\notbf{Simulate $\mathcal{G}$.} We instantiate Algorithm \ref{alg:audit} for our experiments, shown in Figure \ref{fig:tuning_dis_game}. Each execution of $\mathcal{P}$ in $\mathcal{G}$ is an execution of our tuning protocol $\mathcal{H}(\hat{X}, \Omega, \xi, g)$. $\Omega$ contains many base algorithms (DP-SGD instances with different hyper-parameter setups) satisfying the same privacy parameter. $\xi$ is the TNB distribution \cite{papernot2021hyperparameter} shown in Appendix \ref{app:TNBD}. $g$ is the score function.

\notbf{Conclude the lower bound.} Our \textit{null} and \textit{alternative} hypothesis are 
\equienvs
\begin{equation}\label{equ:hypo_test}
\begin{aligned}
\mathbf{H_0}:\text{$X$ was used},\text{\quad}\mathbf{H_1}:\text{$X'$ was used}. 
\end{aligned}
\end{equation}
\equienve
After many simulations of $\mathcal{G}$ where each one gives an assertion for the above hypothesis testing problem, the $\mathrm{FP}$ and $\mathrm{FN}$ are computed by the Clopper-Pearson method \cite{clopper1934use} with a $95\%$ confidence. We then leverage methods proposed in \cite{nasr2023tight} to compute the empirical privacy lower bound $\varepsilon_L^{(X,X',\mathcal{R})}$. We provide the detailed procedure for deriving $\varepsilon_L^{(X,X',\mathcal{R})}$ in Appendix \ref{app:exp_detail}. We omit the notation $(X,X',\mathcal{R})$ under clear context.

\subsection{Audit Scenario Formulation}
\begin{figure}[!t] 
    \centering
    {
    \includegraphics[width=.7\linewidth]{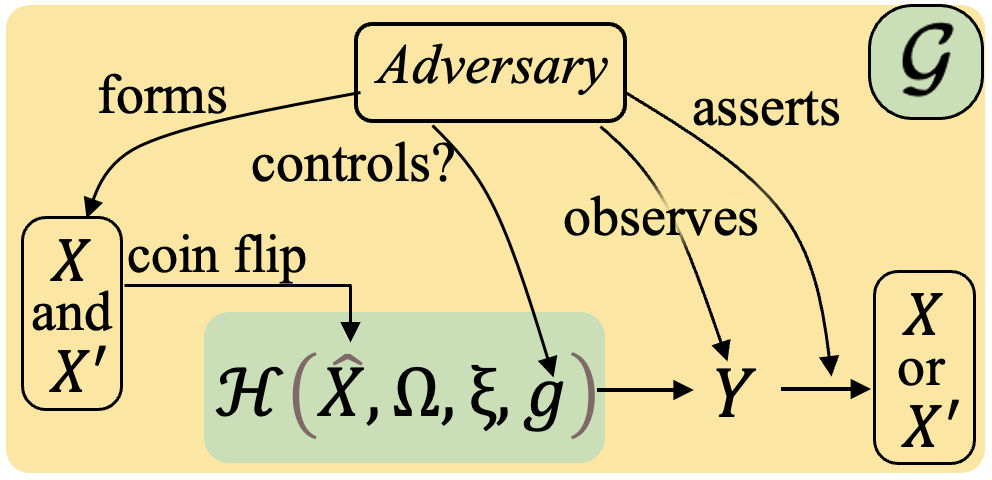}
    }
    \vspace{\fmgftoc}
    \caption{
    Diagram of the distinguishing game $\mathcal{G}$.
   }
    \label{fig:tuning_dis_game} 
    \vspace{\fmgctom}
\end{figure}

This section is to elucidate the \underline{four ``arrows''} originating from the adversary shown in Figure \ref{fig:tuning_dis_game}. 

\notbf{Forming $X,X'$.} W.o.l.g., we assume $X'=X\cup\{\mathbf{z}\}$. Note that the adversary can set $X$ to be any available datasets. $\mathbf{z}$, known as ``canaries'' \cite{nasr2023tight}, is instantiated as follows.
\begin{itemize}
    \item \textit{Weaker version}. The adversary can select $\mathbf{z}$ to be any real-world data, and to have higher distinguishing performance, $\mathbf{z}$ is set to be sampled from a distribution different from those in $X$.
    \item \textit{Stonger version}. The adversary can directly control the gradient of $\mathbf{z}$, {\it a.k.a.}, gradient canary. Specifically, it is assumed that adversary generates $\mathbf{z}= \mathbf{Z}_D$ such that its gradient is a \textit{Dirac} vector $\nabla_w\ell(w;\mathbf{Z}_D)=[C,0,0,\cdots,0]^T$ \cite{nasr2023tight},  i.e., only the first coordinate is $C$.
    
\end{itemize}

\notbf{Score function $g$.} The best model is selected if it has the \textit{highest} score. \textit{This new factor distinguishes auditing $\mathcal{H}$ from all previous auditing tasks}. Formalizing this factor and making corresponding assertions are our key contributions. We formalize two types of adversaries that are only possible. 
\begin{itemize}
    \item \textit{Weaker version}. $g$ \textit{is not} manipulated, e.g., $g$ is a normal routine to evaluate the model's accuracy/loss on an untampered validation dataset.
    \item \textit{Stronger version}. The adversary can arbitrarily control $g$, e.g., $g$ can be a routine to evaluate the model's performance on some malicious dataset.  
    \RevisionChangeStart
    In practice, we believe the score function is some pre-defined function (e.g., the validation accuracy) and is not able to be manipulated. We enforce this setup is to explore the worst-case privacy leakage.
    \RevisionChangeEnd
\end{itemize}

\notbf{Adversary's observation $Y$.} Under the assumption of DP-SGD protocol, the whole training trajectory $\{p_i\}_{i=1}^{N}$ is released. Equivalently, all the checkpoints $\{ w_i\}_{i=1}^{N}$ of the neural network are trivially derivable as each checkpoint is just post-processing of the private gradient. Hence, we can denote the observation as $Y=\{p_1,p_2,\cdots\,p_N, w_1, \cdots, w_N\}$. This information corresponds to line \ref{alg:audit_obs} of Algorithm \ref{alg:audit} or the output of $\mathcal{H}$. Note that including $w_i, i\in\{1,2,\cdots, N\}$ in $Y$ may be redundant; however, it is for notation convenience as we will later refer to the $w_i$ information contained in $Y$.

\begin{figure*}[!ht] 
    \centering
    \subfloat[{$X = \mathbb{F}, \mathbf{z} =  \mathbb{M}[0],g=\textbf{SF}$[1]}]
    {\includegraphics[width=.32\linewidth]{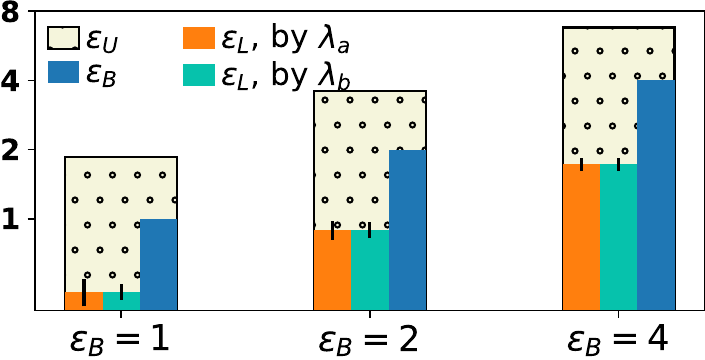}\label{fig:ntnv_1}
    }
    \subfloat[{$X = \mathbb{C}, \mathbf{z} =  \mathbb{S}[0],g=\textbf{SF}$[1]}]
    {\includegraphics[width=.32\linewidth]{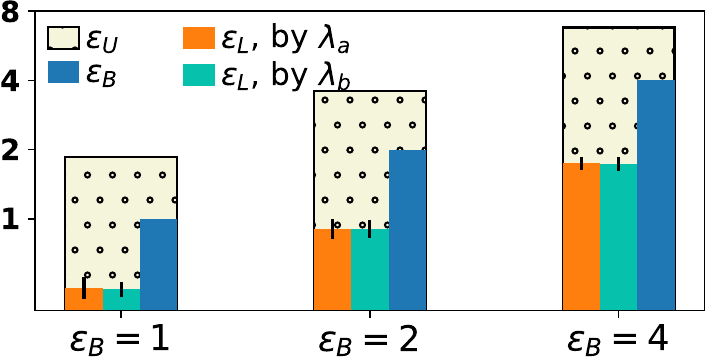}\label{fig:ntnv_2} }
    \subfloat[{$X = \mathbb{S}, \mathbf{z} =  \mathbb{C}[0],g=\textbf{SF}$[1]}]
    {\includegraphics[width=.32\linewidth]{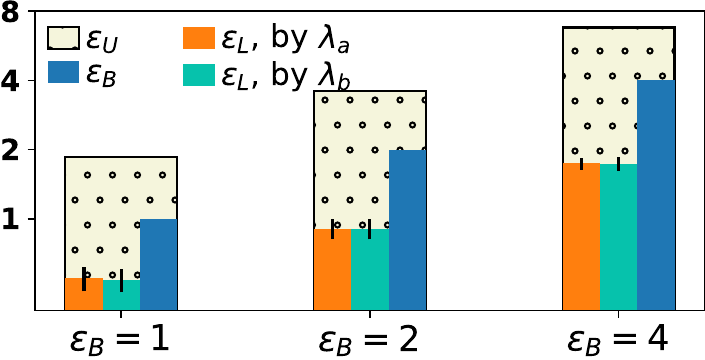}\label{fig:ntnv_3} }\\
    \subfloat[{$X = \mathbb{F}, \mathbf{z} =  \mathbf{Z}_D,g=\textbf{SF}$[1]}]
    {\includegraphics[width=.32\linewidth]{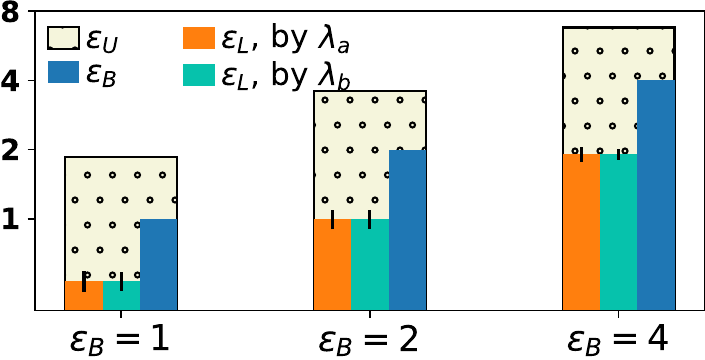}\label{fig:ntnv_4} }
    \subfloat[{$X = \mathbb{C}, \mathbf{z} =  \mathbf{Z}_D,g=\textbf{SF}$[1]}]
    {\includegraphics[width=.32\linewidth]{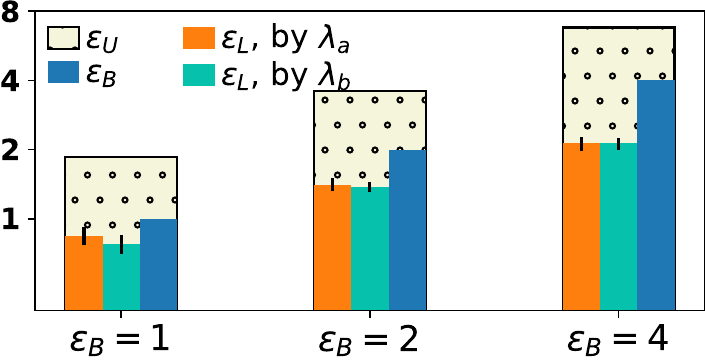}\label{fig:ntnv_5} }
    \subfloat[{$X = \mathbb{S}, \mathbf{z} =  \mathbf{Z}_D,g=\textbf{SF}$[1]}]
    {\includegraphics[width=.32\linewidth]{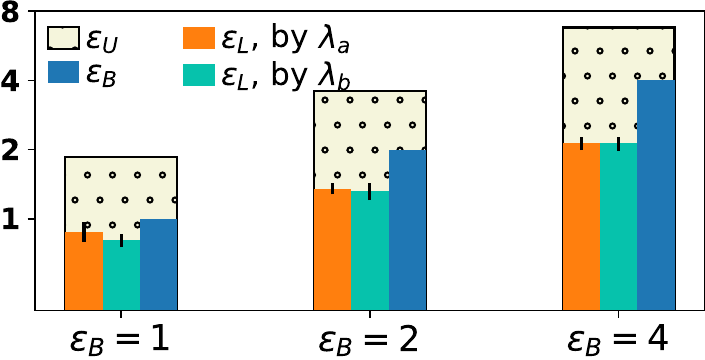}\label{fig:ntnv_6} }\\
    \vspace{\fmgftoc}
    \caption{\textit{\textbf{NTNV}} setup. Rows differ in differing data $\mathbf{z}$; columns differ in training datasets $X$. The vertical axis shows the values for $\varepsilon_U$ and audited $\varepsilon_L$ based on different proxies. Notations are explained by Equation \eqref{equ:notaion_equ}.
    }
    \label{fig:ntnv} 
    \vspace{\fmgctom}
\end{figure*}

\notbf{Adversary's assertion.} Adversary's assertion is exactly the action shown in line \ref{alg:audit_make_assertion} in Algorithm \ref{alg:audit}. This requires the adversary to transform observations $Y$ into binary guesses. The general procedure is as follows. 

The adversary forms a real-number proxy based on observations and compares it to some threshold to make assertions. Proxies are described as follows:

\RevisionChangeStart
A base proxy $\lambda_a$ will be formed following previous work \cite{nasr2021adversary, nasr2023tight} as follows. Compute $\mathbf{z}$'s gradient at each iteration before model update: 
\equienvs
\begin{equation}\label{equ:proxies_a}
\begin{aligned}
    \lambda_a=\frac{1}{N}\sum_{i=1}^{N}\frac{1}{C^2}\langle p_i^\mathbf{z}, p_i\rangle,
\end{aligned}
\end{equation}
\equienve
where $\langle a, b\rangle$ is the inner product. By design, other data examples are independent of $\mathbf{z}$. Hence, we expect $p_i^\mathbf{z}$ is likely to be (approximately) orthogonal to other data's gradient. Moreover, the adversary has access to the score function. Therefore, it seems reasonable to leverage such additional information. To this end, we will also form another proxy $\lambda_b$ based on the score function as follows.
\equienvs
\begin{equation}\label{equ:proxies_b}
\begin{aligned}
    \lambda_b= \lambda_a-g(Y)
\end{aligned}
\end{equation}
\equienve
$\lambda_b$ is our \textit{newly} formed proxy and can be seen as the enhanced version of $\lambda_a$ in auditing hyper-parameter tuning because it tries to include additional information from the score function $g$. 

\RevisionChangeEnd

\notbf{Summary.} We form the following scenarios with increasing levels of threat.
\begin{itemize}
    \item \textit{Normal training and normal validation} (\textit{\textbf{NTNV}}). The training dataset is a natural, normal dataset, and the training model's validation is also normal, i.e., score function $g$ \textit{is not} manipulated. 
    \item \textit{Normal training and controlled validation} (\textit{\textbf{NTCV}}). The training dataset is the same as that of \textit{\textbf{NTNV}}; however, the validation for the trained model is controlled, i.e., score function $g$ \textit{is} manipulated. 
    \item \textit{Empty training and controlled validation} (\textit{\textbf{ETCV}}). The training dataset is empty (malicious), $g$ is the same as that of \textit{\textbf{NTCV}}.
\end{itemize}

\subsection{Evaluation Methods}\label{sec:presentation_eval}
Given the complexity of this subject, here we describe how to evaluate our experimental result.
The used dataset and our code link are provided in Appendix \ref{app:exp_detail}. For notation convenience, we use abbreviations for the used datasets: $\mathbb{F}$ stands for the \underline{F}ASHION dataset, $\mathbb{M}$ for \underline{M}NIST,  $\mathbb{C}$ for \underline{C}IFAR10 and $\mathbb{S}$ for \underline{S}VHN. We use ``[]'' to fetch the information from some data container. For instance, we use $v[0]$ to denote fetching the first coordinate of $v$ if $v$ is a vector. We also abuse the notation and use $Y[w_N]$ to denote fetching the parameter $w_N$ from output/observation $Y$. 

\notbf{Results indexing}. Our main audit results are presented in figures, and we index them in the following form:
\begin{equation}\label{equ:notaion_equ}
    X = \mathbb{F}, \mathbf{z} = \mathbb{M}[I], g=\text{\textbf{SF}}[1],
\end{equation}

which means that such a result corresponds to 1) setting $X$ to be the \underline{F}ASHION dataset; 2) setting the differing data $\mathbf{z}$ to be the $I$-th data sample from \underline{M}NIST dataset; 3) setting the score function $g$ to be the first candidate shown in Table \ref{tab:score_func_setup}. Note that $X'=X\cup \{\mathbf{z}\}$ and we always shuffle the dataset initially.

\begin{table}[!ht] 
\centering
\resizebox{.9\columnwidth}{!}{

\begin{tabular}{ll}
    \toprule
    
    Not manipulated
    &
    \makecell[l]{
    1: $g(Y)=-\sum_i\ell(Y[w_N];\mathbb{V}[i])$\\
    \hspace{0.3cm}$\mathbb{V}$ is original validation dataset
    }
    \\
    \midrule
    
    Manipulated
    &
    \makecell[l]{
    2: $g(Y)=-\ell(Y[w_N];\mathbf{z})$ \\
    3: $g(Y)=(Y[w_0]-Y[w_N])[0]$
    }
    \\
    \bottomrule
\end{tabular}
}

\caption{\underline{S}core \underline{f}unctions are indexed by $\text{\textbf{SF}}[a], a=1,2,3$. }
\label{tab:score_func_setup}
\end{table}

\textit{Score function design consideration.} As it will become clearer in later sections, the rationale behind manipulating the score function to be \textbf{SF}[2] is to expect the training to memorize (having low loss) the different data $\mathbf{z}$, and the best model is selected based on this metric. Manipulating the score function to be \textbf{SF}[3] builds on the fact that for $\mathbf{Z}_D$, it suffices only to investigate the first coordinate of the model to recover any trace of $\mathbf{z}=\mathbf{Z}_D$.

\notbf{Evaluation method.} Our main focus is to compare the following bounds; hence, understanding their intuitive interpretations is beneficial. 
\begin{itemize}
    \item $\varepsilon_L$ is the amount of information leakage the adversary can extract based on the execution of $\mathcal{H}$.
    \item $\varepsilon_B$ is the maximal information leakage due to a single run of the \underline{b}ase algorithm, as guaranteed by theoretical analysis \cite{rdp_subgaussian_mir,abadi2016deep}.
    \item $\varepsilon_U$ is the maximal information leakage due to execution of $\mathcal{H}$, as guaranteed by theoretical analysis \cite{liu2019private,papernot2021hyperparameter}.
\end{itemize}

These bounds are all based on fixed $\sigma$ values. Specifically, after $\sigma$ is fixed for the base algorithm $\mathcal{M}$, we 1) compute $\varepsilon_B$ by previous privacy analysis for DP-SGD such as TensorFlow privacy \cite{tfp}; 2) compute $\varepsilon_U$ for $\mathcal{H}$ by current generic bound for hyper-parameter tuning \cite{papernot2021hyperparameter}; 3) apply privacy audit to $\mathcal{H}$, obtaining $\varepsilon_L$ as shown in Section \ref{sec:game_formulation}. We know that $\varepsilon_B\leq\varepsilon_U$ is always true, and it is interesting to make the following comparison.

\begin{itemize} 
    \item $\varepsilon_L$ V.S. $\varepsilon_U$. This is the main focus. The question to be answered in this comparison is: does hyper-parameter tuning $\mathcal{H}$ practically leak sensitive information ($\varepsilon_L$) as predicted by the current generic bound \cite{papernot2021hyperparameter} ($\varepsilon_U$)?
    \item $\varepsilon_L$ V.S. $\varepsilon_B$. This is another interesting comparison. The question to be answered in this comparison is: How does running a DP-SGD many times and then returning the best (an execution of $\mathcal{H}$) practically leak information ($\varepsilon_L$) compared to a single run of DP-SGD ($\varepsilon_B$)? 
\end{itemize}

\begin{figure*}[!ht] 
    \centering
    \subfloat[{$X = \mathbb{F}, \mathbf{z} =  \mathbb{M}[0],g=\textbf{SF}$[2]}]
    {\includegraphics[width=.32\linewidth]{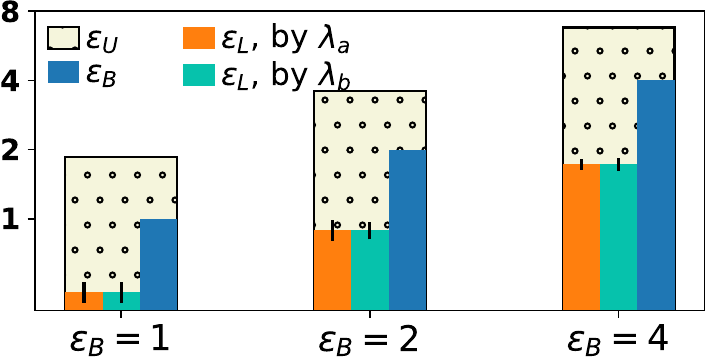}\label{fig:ntcv_1}
    }
    \subfloat[{$X = \mathbb{C}, \mathbf{z} =  \mathbb{S}[0],g=\textbf{SF}$[2]}]
    {\includegraphics[width=.32\linewidth]{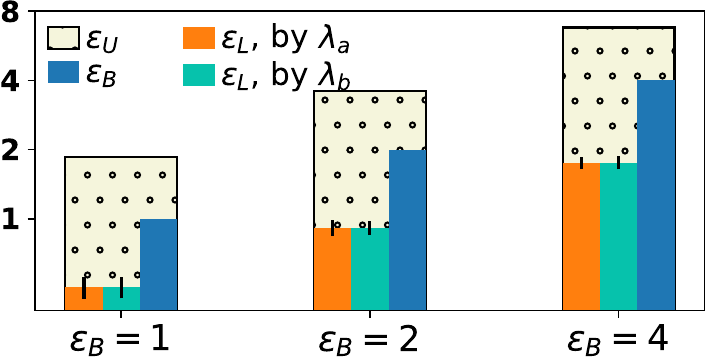}\label{fig:ntcv_2} }
    \subfloat[{$X = \mathbb{S}, \mathbf{z} =  \mathbb{C}[0],g=\textbf{SF}$[2]}]
    {\includegraphics[width=.32\linewidth]{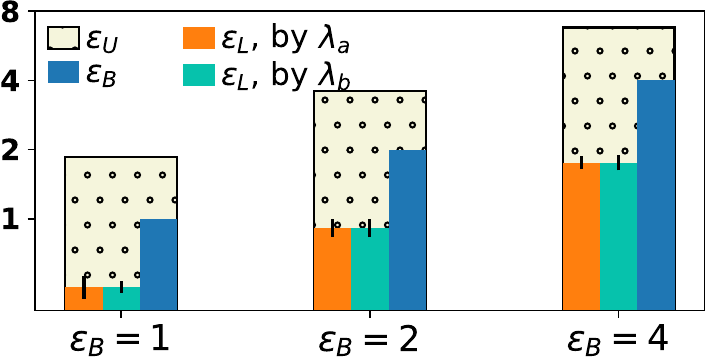}\label{fig:ntcv_3} }\\
    \subfloat[{$X = \mathbb{F}, \mathbf{z} =  \mathbf{Z}_D,g=\textbf{SF}$[3]}]
    {\includegraphics[width=.32\linewidth]{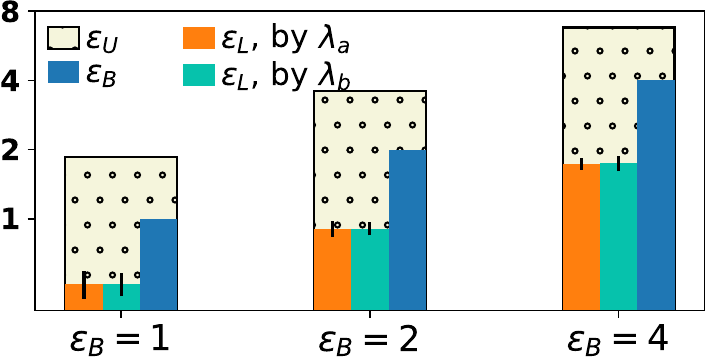}\label{fig:ntcv_4} }
    \subfloat[{$X = \mathbb{C}, \mathbf{z} =  \mathbf{Z}_D,g=\textbf{SF}$[3]}]
    {\includegraphics[width=.32\linewidth]{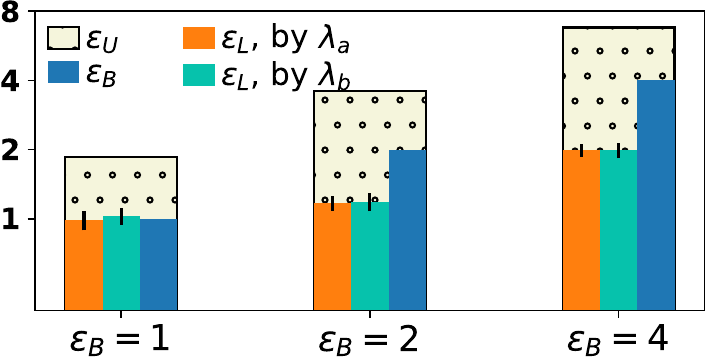}\label{fig:ntcv_5} }
    \subfloat[{$X = \mathbb{S}, \mathbf{z} =  \mathbf{Z}_D,g=\textbf{SF}$[3]}]
    {\includegraphics[width=.32\linewidth]{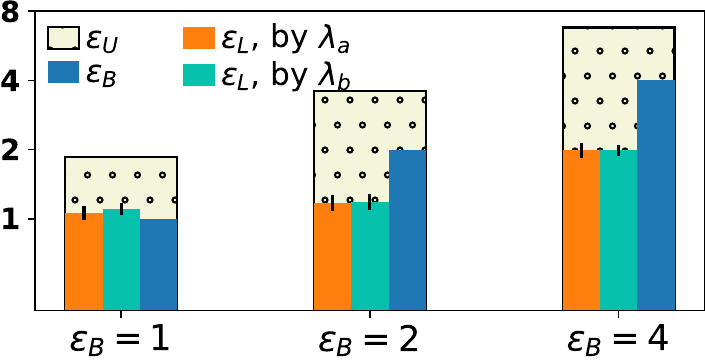}\label{fig:ntcv_6} }\\
    \vspace{\fmgftoc}
    \caption{\textit{\textbf{NTCV}} setup. Sub-figure arrangement is identical to Figure \ref{fig:ntnv}.}
    \label{fig:ntcv} 
    \vspace{\fmgctom}
\end{figure*}

\subsection{Experiments When \texorpdfstring{$g$}{g} Not Manipulated}\label{sec:main_exp_g_not_manipulated}
\notbf{\underline{\textit{\textbf{NTNV}}}.} 
This scenario corresponds to the most practical setup in our experiments. 
Experimental results are shown in Figure \ref{fig:ntnv}, notated according to Section \ref{sec:presentation_eval}. 

\textbf{\textit{Assertion intuition}}. The selection behaves normally, i.e., the best model is selected if it has the highest score (lowest loss) on the original validation dataset. 

By design, higher value of $\lambda_a$ or $\lambda_b$ incentivizes the adversary to accept $\mathbf{H_1}$. The rationale behind these setups is to expect the abnormally differing data (if $X'$ is used, or $\mathbf{z}$ was included in the training) to have a detrimental impact on the training so that the model has a higher loss (lower value of $g(Y)$), making it more distinguishable if $X'$ is used ($\mathbf{z}$ was included in the training).

\textbf{\textit{Results}}. Experimental results are presented in Figure \ref{fig:ntnv}, where we present audited $\varepsilon_L$ results corresponding to proxy $\lambda_a$ or $\lambda_b$. We also present the theoretical upper bound $\varepsilon_U$ for comparison. An obvious phenomenon is that the audited $\varepsilon_L<\varepsilon_B<\varepsilon_U$ across all setups shown in the first row of Figure \ref{fig:ntnv}. The interesting phenomenon is that $\varepsilon_L<\varepsilon_B$ and the gaps between them are obvious. 

In contrast, in Figure \ref{fig:ntnv_5} and Figure \ref{fig:ntnv_6}, when the base algorithm's privacy budget $\varepsilon_B=1$, we see that $\varepsilon_L$ is much closer to $\varepsilon_B$. This confirms that the differing data $\mathbf{z}$ that has \textit{Dirac gradient} gains the adversary more distinguishing power than some natural data. Another phenomenon is that the audited $\varepsilon_L$ under $\varepsilon_B=1$ in Figure \ref{fig:ntnv_4} is weaker than that in Figure \ref{fig:ntnv_5} and Figure \ref{fig:ntnv_6}, this suggests that auditing performance depends on $X$. 

\textbf{\textit{Short summary}.} The adversary cannot even extract more sensitive information than the base algorithm's (a single run of DP-SGD) privacy budget allows, i.e., $\varepsilon_L<\varepsilon_B<\varepsilon_U$. This shows the adversary's power is heavily limited under the most practical setting.

\subsection{Experiments When \texorpdfstring{$g$}{g} Manipulated}\label{sec:main_exp_g_manipulated}
\notbf{\underline{\textit{\textbf{NTCV}}}.} 
This scenario corresponds to some middle-level adversary's power. Experimental results are shown in Figure \ref{fig:ntcv}, notated according to Section \ref{sec:presentation_eval}. The score function \textit{is} manipulated, different from that in \textbf{\textit{NTNV}}.

\textbf{\textit{Assertion intuition}}. The proxy $\lambda_a$ is identical to that in Equation \eqref{equ:proxies_a}, however, $\lambda_b= \lambda_a+g(Y)$ is set in \textbf{\textit{NTCV}}, which is different from that in \textbf{\textit{NTNV}}. This is because $g$ is manipulated. Under the same design considerations, a higher value of $\lambda_a$ or $\lambda_b$ incentivizes the adversary to accept $\mathbf{H_1}$.

\textbf{\textit{Results}}. Experimental results are presented in Figure \ref{fig:ntcv}, organized similarly to Figure \ref{fig:ntnv}. We observe a phenomenon similar to \textbf{\textit{NTNV}} that $\varepsilon_L$ sees a big gap to $\varepsilon_B$ shown in the first row of Figure \ref{fig:ntcv}. In contrast, for the results seen in the second row of Figure \ref{fig:ntcv_5} and Figure \ref{fig:ntcv_6}, when base algorithm's privacy budget $\varepsilon_B=1$, we have $\varepsilon_L\approx\varepsilon_B$. Again, this confirms the \textit{Dirac gradient} canary is more powerful. 

\textbf{\textit{Short summary.}} $\lambda_a$ and $\lambda_b$ have almost the same performance, similar to \textbf{\textit{NTNV}} where $g$ is not manipulated.

\begin{figure}[!t] 
    \centering
    \subfloat[{$X = \emptyset, \mathbf{z} =  \mathbb{M}[0],g=\textbf{SF}$[2]}]
    {\includegraphics[width=.64\linewidth]{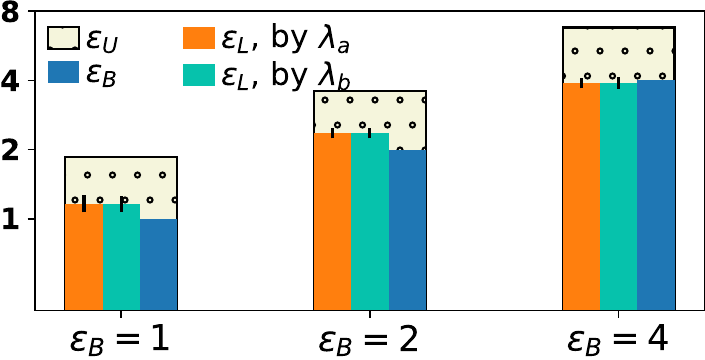}\label{fig:etcv_1}
    }\\
    \subfloat[{$X = \emptyset, \mathbf{z} =  \mathbf{Z}_D,g=\textbf{SF}$[3]}]{\includegraphics[width=.64\linewidth]{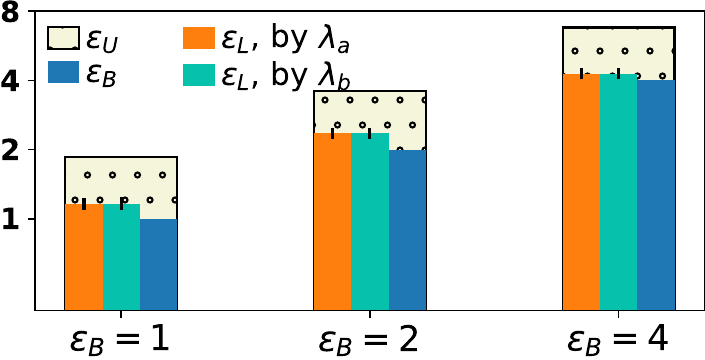}\label{fig:etcv_2} }
    \vspace{\fmgftoc}
    \caption{\textit{\textbf{ETCV}} setup. $\varepsilon_L\geq\varepsilon_B$ is almost true.}
    \label{fig:etcv} 
\end{figure}

\notbf{\underline{\textit{\textbf{ETCV}}}.} 
This scenario corresponds to the greatest adversary's power in our settings. The training dataset is set to be empty, and the score function is manipulated.

\textbf{\textit{Assertion intuition}}. By design, the rationale behind the empty dataset setup is to eliminate the uncertainties due to normal training data's gradient so that audit performance is maximized, as the adversary only cares about the causal effect from $\mathbf{z}$ to the output \cite{DBLP:conf/sp/TschantzSD20}. $\lambda_a, \lambda_b$ are set identically to \textbf{\textit{NTCV}}. Again, higher value of $\lambda_a$ or $\lambda_b$ incentivize the adversary to accept $\mathbf{H_1}$.

\textbf{\textit{Results}}. Experimental results are shown in Figure \ref{fig:etcv}, notated according to Section \ref{sec:presentation_eval}. In Figure \ref{fig:etcv_1}, we can see that $\varepsilon_L\geq\varepsilon_B$ under almost all setups; we also notice that $\varepsilon_L$ still sees a gap to $\varepsilon_U$ under some setups; however, $\varepsilon_L$ gets much closer to $\varepsilon_U$ compared with that in   \textit{\textbf{NTNV}} and \textbf{\textit{NTCV}}. The increased audit performance is due to $X=\emptyset$, which eliminates unwanted disturbances for the adversary. In Figure \ref{fig:etcv_2}, when $\mathbf{z}=\mathbf{z}_D$ is the \textit{Dirac gradient} canary instead of some natural data, we observe $\varepsilon_L\geq\varepsilon_B$ under all setups. 

The above results suggest that, operationally, hyper-parameter tuning does leak additional privacy beyond what's allowed to be disclosed by the base algorithm. This also means that tuning hyper-parameters while only accounting the privacy cost for a single run (i.e., naively taking $\varepsilon_U=\varepsilon_B$ ) is problematic in a rigorous manner. On the other hand, like the results in the previous two setups, we also observe that 1) $\lambda_a$ and $\lambda_b$ have almost the same performance, and 2) there is a big gap between $\varepsilon_L$ and $\varepsilon_U$. 

\textbf{\textit{Short summary.}} Worst-case $X$ setup does bring additional help to the adversary ($\varepsilon_L\approx\varepsilon_B$), but the privacy lower bound derived by audit still sees a gap to the privacy upper bound derived by previous work \cite{papernot2021hyperparameter} ($\varepsilon_L<\varepsilon_U$).

\subsection{Phenomenon Highlight}\label{sec:exp_dis}

\notbf{New finding.} Our \textbf{\textit{NTNV}} and \textbf{\textit{NTCV}} settings give a result that $\varepsilon_L<\varepsilon_U$, leading to the conclusion that weak audit leads to weak lower bound. Although our audit problem differs from the previous audit on DP-SGD's privacy, a weak audit leading to a weak lower bound is also observed previously. However, for \textbf{\textit{ETCV}} setting (strong audit setting), \textbf{what we find is starkly different from the previous audit on DP-SGD}: 
\begin{quote}
    \textit{previous work \cite{nasr2021adversary} shows that maliciously chosen dataset $X$ leads to $\varepsilon_L\approx\varepsilon_U$ (tight result) for auditing DP-SGD's privacy. In contrast, our audit on hyper-parameter tuning shows that $\varepsilon_L$ still sees a noticeable gap to $\varepsilon_U$ even when $X$ is adversarily chosen.}
\end{quote}
This new finding in auditing private hyper-parameter tuning motivates our theoretical study presented in the next section.

\begin{table}[!ht] 
\large
\centering

\resizebox{1\columnwidth}{0.6\height}{
\begin{tabular}{ccc | ccc | ccc | ccc}
\toprule

Proxy
&
$\varepsilon_B$
&
$\varepsilon_U$
&
\multicolumn{9}{c}{$\varepsilon_L$}
\\
\midrule

&
&
&
\multicolumn{3}{c|}{\textbf{\textit{NTNV}} $@$ $X=$}
& 
\multicolumn{3}{c|}{\textbf{\textit{NTCV}} $@$ $X=$}
&
\multicolumn{3}{c}{\textbf{\textit{ETCV}} $@$ $X=$}
\\
&&&&&&&&& \\ [-8pt]

&
&
&
$\mathbb{F}$
&
$\mathbb{C}$
&
$\mathbb{S}$
&
$\mathbb{F}$
&
$\mathbb{C}$
&
$\mathbb{S}$
&
\multicolumn{3}{c}{
$\emptyset$
}
\\
&&&&&&&&& \\ [-6pt]


\makecell[c]{
$\lambda_a$\\
$\lambda_b$
}
&
1
&
1.86
&
\makecell[c]{
$0.54$\\
$0.54$\\
}
&
\makecell[c]{
$0.84$\\
$0.77$\\
}
&
\makecell[c]{
$0.88$\\
$0.81$\\
}
&
\makecell[c]{
$0.52$\\
$0.52$\\
}
&
\makecell[c]{
$0.99$\\
$1.03$\\
}
&
\makecell[c]{
$1.06$\\
$1.11$\\
}
&
\multicolumn{3}{c}{
\makecell[c]{
$1.17$\\
$1.19$\\
}
}
\\
\midrule
\makecell[c]{
$\lambda_a$\\
$\lambda_b$
}
&
2
&
3.59
&
\makecell[c]{
$0.99$\\
$1.00$\\
}
&
\makecell[c]{
$1.41$\\
$1.37$\\
}
&
\makecell[c]{
$1.35$\\
$1.32$\\
}
&
\makecell[c]{
$0.91$\\
$0.91$\\
}
&
\makecell[c]{
$1.17$\\
$1.19$\\
}
&
\makecell[c]{
$1.17$\\
$1.19$\\
}
&
\multicolumn{3}{c}{
\makecell[c]{
$2.36$\\
$2.09$\\
}
}
\\
\midrule
\makecell[c]{
$\lambda_a$\\
$\lambda_b$
}
&
4
&
6.79
&
\makecell[c]{
$1.91$\\
$1.91$\\
}
&
\makecell[c]{
$2.13$\\
$2.13$\\
}
&
\makecell[c]{
$2.13$\\
$2.13$\\
}
&
\makecell[c]{
$1.74$\\
$1.74$\\
}
&
\makecell[c]{
$1.99$\\
$1.98$\\
}
&
\makecell[c]{
$1.99$\\
$1.98$\\
}
&
\multicolumn{3}{c}{
\makecell[c]{
$4.28$\\
$4.68$\\
}
}
\\

\bottomrule
\end{tabular}

}

\caption{
\RevisionChangeStart
Results summary for privacy audit result $\varepsilon_L$ (averaged over 5 runs) obtained under two different proxies when the canary $\mathbf{z}=\mathbf{z}_D$. We also highlight the base algorithm's upper bound $\varepsilon_B$ and upper bound for hyper-parameter tuning using \cite{papernot2021hyperparameter}'s method.
\RevisionChangeEnd
}
\label{tab:two_proxy}
\vspace{-0.4cm}
\end{table}

\RevisionChangeStart

\notbf{How does the score function (selection) leak privacy?} This is a key question. Unlike prior auditing works \cite{nasr2021adversary,jagielski2020auditing,nasr2023tight}, it has not been studied. Table~\ref{tab:two_proxy} summarizes key audit results. First, $\lambda_a$ and $\lambda_b$ perform similarly across all settings. Second, comparing \textbf{\textit{NTNV}} and \textbf{\textit{NTCV}} shows that manipulating the score function offers no clear advantage to the adversary.

This may seem surprising. One might expect that tailoring the metric to pick the best candidate would make distinguishing easier. But our results show otherwise. This is because $\lambda_a$ already reflects the effect of the score function—it is computed from the selected output. So $\lambda_b$ adds no new information.

In fact, the score function output is post-processing of a DP algorithm, and thus remains differentially private. Whether or not $\mathbf{z}$ is included, the induced score distributions are close—just as DP guarantees. So no ``magic'' advantage exists, regardless of the score function, if it’s independent of sensitive data.

\notbf{Is it safe to tune hyper-parameter while only assuming a single run's privacy cost?} Our weakest audit setup, which mimics practical use, suggests yes. Tuning adds little to no extra leakage beyond the base algorithm. This supports the common—but non-rigorous—practice of ignoring tuning overhead before private selection methods were formalized \cite{liu2019private, papernot2021hyperparameter}.

However, under our strongest audit, tuning does leak more privacy. This confirms the importance of formally accounting for the tuning step, justifying both the previous work and our improved bound in Section~\ref{sec:improved_bound}.

\notbf{Can we get a higher lower bound on any other datasets than what we get in \textbf{\it ETCV} setting?} The \textbf{\it ETCV} setup already achieves dataset-independent lower bounds. Since the adversary can inject arbitrary gradient canaries, any bound achievable on real data is also achievable in this setting.

\RevisionChangeEnd

\section{Improved Theoretical Results}\label{sec:improved_bound}
\RevisionChangeStart
In the previous empirical study, a conspicuous gap exists between $\varepsilon_U$ derived by \cite{papernot2021hyperparameter} and $\varepsilon_L$, this makes it interesting to investigate the reason behind it.
\RevisionChangeEnd
Our study in the remaining sections shows that such a gap exists for two reasons. 
\begin{enumerate}[leftmargin=*]
    \item  Current generic privacy upper bound is not tight for DP-SGD;
    \item Adversary's power is not strong enough because it is hard for the adversary to instantiate the worst-case score function $g$. 
\end{enumerate}
Regarding 1), we provide improved privacy results and elucidate on the special property of DP-SGD leading to the improvement;
\RevisionChangeStart
Our analysis is generalizable beyond DP-SGD, i.e., as will be shown, our analysis works for any base algorithm that can be expressed within the $f$-DP framework.
\RevisionChangeEnd
For 2), we present meaningful findings about the score function. 

\vspace{0.05cm}
\notbf{Problem modelling}.
Informally, the privacy problem for our private tuning algorithm $\mathcal{H}$ (Algorithm \ref{alg:ps}) can be compactly described as the following optimization formulation.

\begin{equation}\label{equ:privacy_opti}
\begin{aligned}
    \textbf{\textit{minimize:}} & \text{\quad$\varepsilon_\mathcal{H}$}\\
    \textbf{\textit{subject to:}} & \text{\quad $\mathcal{H}$ satisfies $(\varepsilon_\mathcal{H}, \delta_\mathcal{H})$-DP given $\delta_\mathcal{H}$};\\
    &\text{\quad base algorithm's privacy is \ding{70}}
\end{aligned}
\end{equation}
It is self-evident that the tightness of $\varepsilon_\mathcal{H}$ depends on how tight \ding{70} is.  
The critical part is how we represent the privacy guarantee \ding{70}. Under our optimization formulation, previous work describes \ding{70} as follows: 1) Liu et.al \cite{liu2019private} represents the base algorithm by $(\varepsilon,\delta)$-DP; 2) Papernot et al. \cite{papernot2021hyperparameter} does that by $(\alpha,\gamma)$-RDP, obtaining improved results over \cite{liu2019private}. Can we do better? As will be shown below, the answer is yes if we represent the base algorithm's privacy by $f$-DP.

\subsection{Preliminaries: \texorpdfstring{$f$}{f}-DP}

$f$-DP \cite{dong2019gaussian}, a privacy formulation with a finer resolution, reflects the nature of private mechanisms by a \textit{function} \cite{zhu2022optimal} rather than a single pair of parameters. Our improved results are derived based on the $f$-DP framework. We introduce the necessary definitions and technical preliminaries in the following.

\begin{definition}[Trade-off function \cite{dong2019gaussian}]
    For a hypothesis testing problem over two distributions $P,P'$, define the trade-off function as:
    \equienvs
    \begin{equation}\nonumber
        T_{P,P'}(\mathrm{FP})=\inf_{\mathcal{R}}\{\mathrm{FN}_{\mathcal{R}}:\mathrm{FP}_\mathcal{R}\leq \mathrm{FP}\}
    \end{equation}
    \equienve
    where decision rule $\mathcal{R}$ takes input a sample from $P$ or $P'$ and decides which distribution produced that sample. The infimum is taken over all decision rule $\mathcal{R}$.
\end{definition}
The trade-off function governs the best one can achieve when distinguishing $P$ from $P'$, i.e., by the optimal/smallest type II error ($\mathrm{FN}$) at fixed type I error ($\mathrm{FP}$). The optimal $\mathrm{FN}$ is achieved via the likelihood ratio test, which is also known as the fundamental \textit{Neyman–Pearson lemma} \cite{neyman1933ix} (please refer to Appendix \ref{app:np_lemma}). We denote 
\begin{equation}\nonumber
    \text{$g\geq f$ if $g(x)\geq f(x), \forall x\in[0,1]$}.
\end{equation}
\begin{definition}[$f$-DP \cite{dong2019gaussian}]
    Let $f:[0,1]\rightarrow [0,1]$ be a trade-off function. A mechanism $\mathcal{M}$ satisfies f-DP if 
    \equienvs
    \begin{equation}\nonumber
        T_{\mathcal{M}(X),\mathcal{M}(X')}\geq f
    \end{equation}
    \equienve
    for all adjacent dataset pairs $X, X'$
\end{definition}
$\mathcal{M}$ being $f$-DP means that any possible error pair $(\mathrm{FP}, \mathrm{FN})$ resulting from distinguishing $\mathcal{M}(X)$ from $\mathcal{M}(X')$ is lower-bounded by the curve specified by $f$. To see why $(\varepsilon,\delta)$-DP is loose. We must express $(\varepsilon,\delta)$-DP with the language of $f$-DP. This is done via the following proposition.
\begin{proposition}[$(\varepsilon,\delta)$-DP equals to $f_{\varepsilon,\delta}$-DP \cite{dong2019gaussian,wasserman2010statistical}]\label{[prop:eps_delta_f_dp]}
    $\mathcal{M}$ is $(\varepsilon,\delta)$-DP if and only if it is $f_{\varepsilon,\delta}$-DP where the trade-off function $f_{\varepsilon,\delta}$ is {$$ f_{\varepsilon,\delta}(x)=\max{(0,1-\delta-\mathrm{e}^\varepsilon x,\mathrm{e}^{-\varepsilon}(1-\delta-x))}$$}
\end{proposition}

\noindent $f$-DP implies $(\varepsilon,\delta)$-DP and conversion from $f$-DP to $(\varepsilon,\delta)$-DP is via Algorithm \ref{alg:fdp_to_eps_delta} (restatement of Proposition 6 of \cite{dong2019gaussian}).

\begin{figure}[!t] 
    \centering
    {
    \includegraphics[width=.63\linewidth]{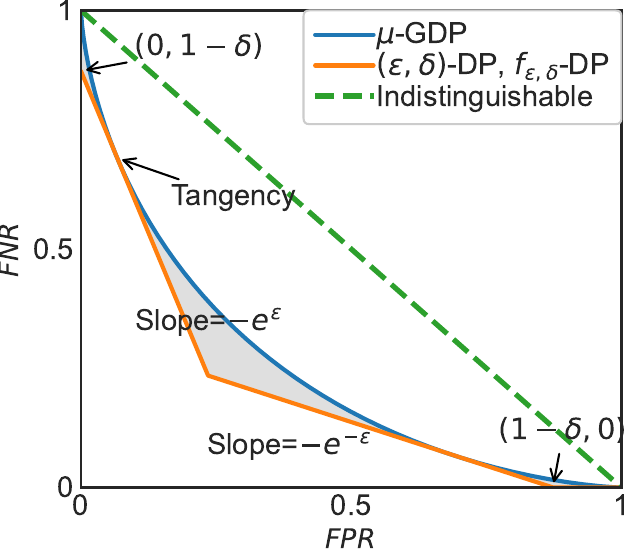}
    }
    \vspace{\fmgftoc}
    \caption{
    For the Gaussian mechanism $\mathcal{M}(X)=q(X)+\mathcal{N}(0,\sigma^2\mathbb{I}^d)$ where the query function $q(X)\in\mathbb{R}^d$ has unit $\ell_2$-sensitivity, it is exactly
    $1/\sigma$-GDP \cite{dong2019gaussian}. It is also some $(\varepsilon,\delta)$-DP; however, $\mu$-GDP characterization saves the shaded area in gray.
    }
    \label{fig:GDP_better} 
\end{figure}

In plain words, $f_{\varepsilon,\delta}$-DP (or $(\varepsilon,\delta)$-DP) for some mechanism $\mathcal{M}$ is the two symmetric straight lines lower-bounding the true/faithful trade-off function of $\mathcal{M}$. This is drawn in Figure \ref{fig:GDP_better}. For the Gaussian mechanism, which is probably the most basic private mechanism, using $(\varepsilon,\delta)$-DP to characterize its privacy is not tight/faithful; in contrast, the following special family of trade-off functions is tight.

\begin{definition}[$\mu$-Gaussian DP ($\mu$-GDP) \cite{dong2019gaussian}]\label{def:gdp}
    The trade-off function of distinguishing $\mathcal{N}(0,1)$ from $\mathcal{N}(\mu,1)$ is
    {$$ G_\mu(x)=T_{\mathcal{N}(0,1),\mathcal{N}(\mu,1)}(x)=\Phi(\Phi^{-1}(1-x)-\mu),$$}where $\Phi$ be the c.d.f. of standard normal distribution. A private mechanism $\mathcal{M}$ satisfies $\mu$-GDP if it is $G_\mu$-DP

\end{definition}
The analytical expression of $\mu$-GDP is due to the application of Neyman–Pearson lemma on distinguishing $\mathcal{N}(0,1)$ from $\mathcal{N}(\mu,1)$ \cite{dong2019gaussian}. $\mathcal{M}$ satisfying $\mu$-GDP means that distinguishing $\mathcal{M}(X)$ from $\mathcal{M}(X')$ is at least as hard as distinguishing $\mathcal{N}(0,1)$ from $\mathcal{N}(\mu,1)$. Figure \ref{fig:GDP_better} explains why $(\varepsilon,\delta)$-DP is loose: $(\varepsilon,\delta)$-DP is \textit{strictly more conservative} than $\mu$-GDP when characterizing the privacy of Gaussian mechanism.

\begin{algorithm}[!ht]
\caption{
$f$-DP to $(\varepsilon,\delta)$-DP \cite{dong2019gaussian} \hspace{0.1cm} 
}\label{alg:fdp_to_eps_delta}

\begin{algorithmic}[1]
\equsize
\renewcommand{\algorithmicrequire}{\textbf{Input:}}
\renewcommand{\algorithmicensure}{\textbf{Output:}}

\Require {$f$, trade-off function; $\delta$, privacy parameter}

\State  If $\delta < 1-f(0)$, return $\infty$
\State Compute $\varepsilon=\inf\{a:f(x)\geq 1-\delta-\mathrm{e}^a x, \forall x\in[0,1]\}$ via binary search

\Ensure $\max\{0, \varepsilon\}$

\end{algorithmic}
\end{algorithm}
The following corollary gives a closed-form solution for optimal/lossless conversion from $\mu$-GDP to $f_{\varepsilon,\delta}$-DP (or $(\varepsilon,\delta)$-DP) in accordance with Algorithm \ref{alg:fdp_to_eps_delta}.

\begin{corollary}[Conversion from $\mu$-GDP to $(\varepsilon,\delta)$-DP formulation \cite{dong2019gaussian,balle2018improving}]\label{cor:from_gdp_to_eps_delta}
    A mechanism is $\mu$-GDP if and only if it is $f_{\varepsilon,\delta(\varepsilon)}$-DP (or $(\varepsilon,\delta(\varepsilon))$-DP) $\forall \varepsilon\geq 0$ where
    \equienvs
    \begin{equation}\label{equ:from_gdp_to_eps_delta}
    \begin{aligned}
        \delta(\varepsilon)=\Phi(-\frac{\varepsilon}{\mu}+\frac{\mu}{2})-\mathrm{e}^\varepsilon\Phi(-\frac{\varepsilon}{\mu}-\frac{\mu}{2})
    \end{aligned}
    \end{equation}
    \equienve
\end{corollary}

\begin{remark}
    The purpose of introducing all previous technical preliminaries (especially Figure \ref{fig:GDP_better}) is not only to necessarily introduce $f$-DP itself but also to understand why we can obtain improvements under $f$-DP framework (see Example \ref{example:why_improvement}).
\end{remark}

\notbf{DP-SGD is asymptotic $\mu$-GDP.} $\mu$-GDP is pivotal because it asymptotically characterizes the privacy of any DP-SGD instance having many compositions of Gaussian mechanisms \cite{dong2019gaussian}:
\begin{corollary}[GDP approximation \cite{dong2019gaussian} for DP-SGD]\label{cor:gdp_approx}
    DP-SGD is asymptotically $\mu$-GDP with
    \equienvs
    \begin{equation}\nonumber
    \begin{aligned}
        \mu=\sqrt{2} \tau\sqrt{N} \cdot \sqrt{\mathrm{e}^{\sigma^{-2}} \cdot \Phi\left(1.5 \sigma^{-1}\right)+3 \Phi\left(-0.5 \sigma^{-1}\right)-2}
    \end{aligned}
    \end{equation}
    \equienve
    where $\sigma=\sigma'/C$ and $\sigma'$ is s.t.d. of the Gaussian noise; $C$ is the clipping threshold; $\tau$ and $N$ is the sampling ratio and number of total iteration of DP-SGD.
\end{corollary}

\subsection{Our Contribution: Improved Results}\label{sec:imp_fdp}
A critical fact about Corollary \ref{cor:gdp_approx} is that computing the exact trade-off function is \#P-hard \cite{dong2019gaussian} (even harder than NP problems), which makes it necessary to resort to approximations if one aims at tighter results within the $f$-DP framework. Specifically, the error (\textit{pointwise error} between the asymptotical GDP trade-off function and the true trade-off function) decays at a rate of $1/\sqrt{N}$ for DP-SGD, shown by \cite{dong2019gaussian}. Therefore, using Corollary \ref{cor:gdp_approx} requires $N$ to be large enough. Such a condition holds for probably most DP-SGD applications, especially for training large models (e.g., $N>10^4$ in \cite{abadi2016deep} and $N>10^5$ in \cite{krizhevsky2017imagenet,vaswani2017attention}). 

Based on all of the above preparations, we are ready to approach our privacy problem by filling in the missing part of Equation \eqref{equ:privacy_opti} based on Corollary \ref{cor:gdp_approx}:

\begin{equation}\label{equ:privacy_opti_gdp}
\begin{aligned}
    \textbf{\textit{minimize:}} & \text{\quad$\varepsilon_\mathcal{H}$}\\
    \textbf{\textit{subject to:}} & \text{\quad $\mathcal{H}$ satisfies $(\varepsilon_\mathcal{H}, \delta_\mathcal{H})$-DP given $\delta_\mathcal{H}$};\\
    &\text{\quad base algorithm's privacy is $\mu$-GDP}
\end{aligned}
\end{equation}
In the following, we first revisit the central question of how selection (the score function) leaks privacy. 

In Section \ref{sec:exp_dis}, we showed that the adversary gains no advantage even if the score function $g$ is maliciously manipulated. A natural question arises: is there any score function that brings more advantage to the adversary?

\notbf{One-to-one mapping $g$ is the worst-case necessarily.} 
The score function $g$ is a function that maps the output of the base algorithm to a real number,
If some $g$ happens to map two distinct inputs to the same score (hence, a randomized tie-breaking will be enforced), how does such $g$ affect the privacy of $\mathcal{H}$ compared to one-to-one mapping score functions? 

Intuitively, such $g$ will only make $\mathcal{H}$ more private as new uncertainty is injected. We can gain more intuition by considering the extreme case: if $g$ only outputs a constant, then $\mathcal{H}$ is just as private as the base algorithm. Our theorem in the following
formalizes such intuition.

\begin{theorem}[Necessary worst-case $g$, proof in Appendix \ref{app:proof_bijective_large}]\label{thm:bijective_large}
Let distribution $P$ be over some finite alphabets $\Gamma$, and define a distribution $F_{k,g}$ as follows. 

First, make $k>0$ independent samples $\{x_1,x_2,\cdots, x_k\}$ 
from $P$; second, output $x_i$ such that the score $g(x_i)$ computed by a score function $g:\Gamma\rightarrow\mathbb{R}$ is the maximum over these samples. Similarly, we define another distribution $P'$ over the same alphabets $\Gamma$ and derive a distribution $F'_{k,q}$ as the counterpart to $F_{k,g}$. 

For any score function $\hat{g}$, which is \textbf{not} a one-to-one mapping (hence a randomized tie-breaking is needed), there always exists a one-to-one mapping $g^*$ satisfying
\equienvs
\begin{equation}\label{equ:bi_large}
\begin{aligned}
    \mathcal{D}_{\alpha}(F_{k, \hat{g}}||F'_{k,\hat{g}}) \leq \mathcal{D}_{\alpha}(F_{k,g^*}||F'_{k,g^*}).
\end{aligned}
\end{equation}
\equienve
Moreover, similar inequality also holds when $k$ follows a general distribution $\xi$.
\end{theorem}

The above result is derived under RDP (Definition \ref{def:RDP}) and it tells us crucial facts: \textit{A score function that induces a strict total order for elements in $\Gamma$ tends to be less private}. 
Thus, a one-to-one mapping is necessary to be the worst case for the score function $g$. 

Note that, in previous work \cite{liu2019private,papernot2021hyperparameter}, the score function $g$ is assumed to be one-to-one mapping \textit{by default for simplicity}. We show that such treatment is valid due to privacy considerations; to our knowledge, the above theorem is the first rigorous proof validating such an assumption.

Theorem \ref{thm:bijective_large} also holds when $\Gamma$ is infinite because R\'enyi divergence can be approximated arbitrarily well by finite partition \cite[Theorem 10]{van2014renyi}. With $g$'s necessary condition determined, we can introduce our improved privacy results.

\notbf{Notation.} Let $y,y'\in \mathcal{Y}$ be the output of the base algorithm (DP-SGD, a single run) corresponding to adjacent input dataset $X,X'$, respectively. Let $P,P'$ be the induced score distribution after the score function $g$ takes input $y,y'$, respectively. With some abuse of notation, we use $P(x), F(x)$ to denote the p.d.f. and c.d.f. for distribution $P$ (similarly, we have $P'(x), F'(x)$ w.r.t. $X'$). Based on the assumption that $g$ is a one-to-one mapping, the selection is essentially among samples from $P$ (or $P'$ if $X'$ is the input). 

Let $Q$ be the distribution of the score of the model outputted by $\mathcal{H}$. Let us for now consider the distribution $\xi$ in $\mathcal{H}$ is a point mass on some $k>0$, i.e., $\operatorname{Pr}(k)=1$. Then, the p.d.f. $Q(x)$ is 
\equienvs
\begin{equation}\label{equ:order_stats_pdf}
\begin{aligned}
    Q(x) = k P(x) (F(x))^{k-1}
\end{aligned}
\end{equation}
\equienve
as well-studied in \textit{order statistics} \cite{david2004order}, i.e., it is the distribution of the maximal sample among $k$ independent draws.

When distribution $\xi$ is some general distribution, define the function
\equienvs
\begin{equation}\label{equ:phi}
\begin{aligned}
    \omega_\xi(x)=\sum_{k\sim\xi}k\cdot\operatorname{Pr}_\xi(k)\cdot x^{k-1}
\end{aligned}
\end{equation}
\equienve
and then $Q$ is a mixture distribution, i.e.,  
\equienvs
\begin{equation}\label{equ:mix_q}
\begin{aligned}
    Q(x) =\sum_{k\sim\xi}\operatorname{Pr}_\xi(k)\cdot k P(x) (F(x))^{k-1}= P(x) \omega_\xi(F(x)).
\end{aligned}
\end{equation}
\equienve
Distribution $Q'$'s p.d.f. corresponding to $X'$ being the input is computed similarly. Now, we are ready to present our improved privacy upper bound.
\begin{theorem}[General form, proof in Appendix \ref{app:proof_of_improved_privacy_upper_bound}]\label{thm:improved_privacy_upper_bound}
Suppose the base algorithm is $f$-DP, then $\mathcal{H}$ is $(\varepsilon_\mathcal{H},\delta_\mathcal{H})$-DP where
\begin{equation}\label{equ:eps_by_fdp}
\begin{aligned}
    \varepsilon_\mathcal{H} = \varepsilon + \max_{a\in[0,1]}\log{{\frac{\omega_\xi(1-a)}{\omega_\xi(b)}}},
\end{aligned}
\end{equation}
where $b =f(a)$ and $\varepsilon$ is computed by Algorithm \ref{alg:fdp_to_eps_delta} whose two input arguments are the trade-off function $f$ and $\delta=\delta_\mathcal{H}/\omega_\xi(1)$ ($\omega_\xi$ is defined in Equation \eqref{equ:phi}).
\end{theorem}
We present our $f$-DP accountant for private selection in Algorithm \ref{alg:fdp_accountant} according to Theorem \ref{thm:improved_privacy_upper_bound}.

\begin{algorithm}[!ht]
\caption{
$f$-DP Accountant for $\mathcal{H}$
}\label{alg:fdp_accountant}

\begin{algorithmic}[1]
\equsize
\renewcommand{\algorithmicrequire}{\textbf{Input:}}
\renewcommand{\algorithmicensure}{\textbf{Output:}}

\Require {trade-off function $f$ s.t. the base algorithm is $f$-DP, $\xi$ distribution of $\mathcal{H}$, $\delta_\mathcal{H}$}

\State $\delta\gets\delta_\mathcal{H}/\omega_\xi(1)$\Comment{$\omega_\xi$ is from Equation \eqref{equ:phi}}
\State $\varepsilon\gets$ input $f$ and $\delta$ to Algorithm \ref{alg:fdp_to_eps_delta}
\State $\varepsilon_\mathcal{H}\gets\varepsilon+\max_{a\in[0,1]}\log{(\omega_\xi(1-a)/\omega_\xi(f(a)))}$
\Ensure $\varepsilon_\mathcal{H}$

\end{algorithmic}
\end{algorithm}

Given that the base algorithm is some $\mu$-GDP, we immediately arrive at the improved result for hyper-parameter tuning by plugging in its specific trade-off function.
\begin{corollary}[Improved result for DP-SGD]\label{cor:improved_result_under_gdp}
    If the base algorithm if $\mu$-GDP (or $G_\mu$-DP), then $\mathcal{H}$ is $(\varepsilon_\mathcal{H},\delta_\mathcal{H})$-DP where
    \begin{equation}\label{equ:eps_by_gdp}
    \begin{aligned}
        \varepsilon_\mathcal{H} = \varepsilon + \max_{a\in[0,1]}\log{{\frac{\omega_\xi(1-a)}{\omega_\xi(G_\mu(a))}}}
    \end{aligned}
    \end{equation}
    with $G_\mu(a)$ is in Definition \ref{def:gdp} 
    and {\small $ \delta_\mathcal{H}/ \omega_\xi(1)=\Phi(-\frac{\varepsilon}{\mu}+\frac{\mu}{2})-\mathrm{e}^\varepsilon\Phi(-\frac{\varepsilon}{\mu}-\frac{\mu}{2})$} determines $\varepsilon$.
\end{corollary}
\notbf{Intuitive explanation of our results.} 
\RevisionChangeStart
The intuition is that the selection (choosing the best of many independent runs) results in a different output distribution than when running the base algorithm only once. And it will deteriorate the final privacy bound, this is shown in Equation \ref{equ:eps_by_fdp}: there is an increase (deteriorating) of the $\varepsilon$ parameter compared to the base algorithm's parameter. How the results deteriorate depends on $\xi$.

Modeling the base algorithm with $f$-DP instead of $(\varepsilon,\delta)$-DP leads to tighter bounds. By the post-processing property \cite{dong2019gaussian}, the score output retains the same $f$-DP as the base algorithm. In Equation~\eqref{equ:eps_by_fdp}, $a$ represents the false negative (FN), and $b = f(a)$ is the optimal false positive (FP) at that FN.

If the base algorithm satisfies $\mu$-GDP, then $b = G_\mu(a)$. But if modeled using $(\varepsilon, \delta)$-DP, we only get $b = f_{\varepsilon,\delta}(a)$. Figure~\ref{fig:GDP_better} shows that $G_\mu(a) \geq f_{\varepsilon,\delta}(a)$. Since $\omega_\xi$ is increasing, this gives a tighter (smaller) $\varepsilon_\mathcal{H}$ when using GDP. The following example illustrates the gain from using $f$-DP.
\RevisionChangeEnd

\begin{example}\label{example:why_improvement}
    Suppose the base algorithm (DP-SGD) satisfies $1$-GDP and $\xi$ is the TNB distribution with parameter $\eta=1, \nu=10^{-2}$ (in this case, $\xi$ is geometric distribution, and we recover the case studied by Liu et al. \cite{liu2019private}). Hence, it allows us to make meaningful comparisons.
    
    For $\delta=10^{-5}$, the base algorithm is also $(4.36, 10^{-5})$-DP or $f_{4.36, 10^{-5}}$-DP. If $b=G_1(a)$ in Equation \eqref{equ:eps_by_fdp}, which is how we represent the base algorithm's privacy, we have $\max_{a\in[0,1]}\log{{\frac{\omega_\xi(1-a)}{\omega_\xi(G_1(a))}}}=3.3$; however, if $b=f_{4.36, 10^{-5}}(a)$, which equals to how the base algorithm is modeled by Liu et al. \cite{liu2019private}, $\max_{a\in[0,1]}\log{{\frac{\omega_\xi(1-a)}{\omega_\xi(f_{4.36, 10^{-5}}(a))}}}=16.5>3.3$ is only what we can derive. Thus, a huge improvement is obtained, and this is due to the saved shaded area in gray shown in Figure \ref{fig:GDP_better}.
\end{example}

\subsection{Significance Statement}\label{sec:bound_disc}

\RevisionChangeStart

\notbf{Our result is generalizable and tighter.}  We observe that 1) \cite{liu2019private} only supports geometric $\xi$, and 2) \cite{papernot2021hyperparameter} only supports truncated negative binomial and Poisson $\xi$. It is unclear how to handle arbitrary $\xi$, and prior results require manual, case-specific analysis. In contrast, our result works for any $\xi$ in protocol $\mathcal{H}$. Computing $\omega_\xi$ is always numerically stable, as $\omega_\xi(x)$ is bounded on $[0,1]$.

As shown in Section~\ref{sec:bound_comparison}, our bound improves over prior work. For example, if $\xi$ always outputs 1 (i.e., run the base algorithm once), Equation~\eqref{equ:eps_by_fdp} gives $\varepsilon_\mathcal{H} = \varepsilon$ and $\delta_\mathcal{H} = \delta$, matching the base guarantee. This shows our result is tight for general $\xi$. In contrast, RDP-based analysis is loose due to lossy conversion to $(\varepsilon, \delta)$-DP \cite{zhu2022optimal}.

\RevisionChangeEnd

\notbf{Extension beyond DP-SGD.} Our above example shows that representing the privacy of the base algorithm with finer resolution (from $(\varepsilon,\delta)$-DP to $f$-DP) leads to improvements in the privacy upper bound. Similar conclusions also hold when switching from RDP \cite{papernot2021hyperparameter} to $f$-DP as RDP is also observed to be lossy within the $f$-DP framework \cite{zhu2022optimal}, i.e., RDP shares the same weakness as that of the $(\varepsilon,\delta)$-DP. 
\RevisionChangeStart
We select DP-SGD as our base algorithm because of its popularity in the literature, but our result is not limited to DP-SGD. In fact, any private base algorithm analyzed by $(\varepsilon,\delta)$-DP or RDP can be represented by $f$-DP with finer resolution. And switching to $f$-DP and using our privacy accountant can also bring improvements. The reason is depicted in Figure \ref{fig:GDP_better}: using $f$-DP avoids the unnecessary region shaded in gray.
\RevisionChangeEnd

\vspace{-0.1cm}
\section{Further Evaluation}\label{sec:further_eval}

\subsection{Stronger Audit via Reduction}\label{sec:reduction_base}
\notbf{Motivation for final audit trial.} In the presence of our improved privacy upper bound, we immediately want to assess its tightness by privacy audit for general $\xi$ such that $\operatorname{Pr}_\xi(1)<1$. This requires we derive reasonably strong lower bounds to be informative. We should avoid ad hoc audit setups for real-world training tasks (Section \ref{sec:exp}). We need to form our audit with theoretical-justified power. 

This section is to serve such a purpose. A part of the design considerations relies on our Theorem \ref{thm:bijective_large} in the last section.

\notbf{1) Base algorithm reduction.} Our threat model will be based on the assumption made by DP, i.e., the adversary knows the membership of all data used to update the model in each iteration except for the differring data $\mathbf{z}$ (the \textit{strong adversary assumption} \cite{cuff2016differential,li2013membership}). 

This means the adversary can always subtract the gradient of other data from $p_i$ in each iteration. Hence, any adjacent dataset pair $X,X'$ is equivalent to $X=\emptyset,X'=\mathbf{z}$ from the adversary's view. This allows us to make two reduction steps for the base algorithm (DP-SGD).

\textit{\ding{182} First reduction} (from Equation \eqref{equ:r1_s} to Equation \eqref{equ:r1_e}). Let $\sigma$ noise s.t.d. shown in Equation \eqref{equ:dpsgd}. Now assume that the sampling ratio $\tau=1$, i.e., full-batch gradient descent. Given $X=\emptyset,X'=\mathbf{z}$, then, at each iteration, for the adversary, the private gradient $p_i$ is as follows.

\equienvs
\begin{equation}\label{equ:r1_s}
\begin{aligned}
    & p_i|X=R_i \sim \mathcal{N}(0, C^2\sigma^2\mathbb{I}^d)\\
    &p_i|{X'}=(\nabla_{\mathbf{z}} + R_i) \sim \mathcal{N}(\nabla_{\mathbf{z}}, C^2\sigma^2\mathbb{I}^d),
\end{aligned}
\end{equation}
\equienve
where $p_i|X$ denotes the random variable conditioned on $X$ was chosen and $\nabla_{\mathbf{z}}= \nabla_w\ell(w_{i-1},z)$ with $\|\nabla_{\mathbf{z}}\|_2=C$ (assume maximal $\ell_2$-norm is reached). 
The adversary \textit{can always construct a rotational matrix} $U_{\mathbf{z}}\in \mathbb{R}^{d\times d}$ such that $ p_i$ can be reduced as follows.
\equienvs
\begin{equation}\label{equ:r1_m}
\begin{aligned}
     U_{\mathbf{z}} p_i|X\sim \mathcal{N}(0, C^2\sigma^2\mathbb{I}^d), \text{\quad} U_{\mathbf{z}} p_i|X'\sim \mathcal{N}(U_{\mathbf{z}}\nabla_\mathbf{z}, C^2\sigma^2\mathbb{I}^d)
\end{aligned}
\end{equation}
\equienve
where $U_{\mathbf{z}}\nabla_\mathbf{z}=[C,0,0,\cdots]^T$. This is because  Gaussian noise with $\sigma^2\mathbb{I}^d$ covariance is rotational invariant, i.e., 
\equienvs
\begin{equation}\nonumber
\begin{aligned}
 &\mathbf{Cov}(U_{\mathbf{z}}R_i)=U_{\mathbf{z}}\mathbf{Cov}(R_i)U_{\mathbf{z}}^T =C^2\sigma^2\mathbb{I}^d = \mathbf{Cov}(R_i)\\
\end{aligned}
\end{equation}
\equienve
where $\mathbf{Cov}(R_i)$ is the covariance matrix of Gaussian random vector $R_i$. After the rotation, for the adversary, 
only the first coordinate carries useful information about $\mathbf{z}$. 

Because a noise vector $R_i\sim \mathcal{N}(0, C^2\sigma^2\mathbb{I}^d)$ and its rotated version $U_{\mathbf{z}}R_i\sim\mathcal{N}(0, C^2\sigma^2\mathbb{I}^d)$ possessing $\sigma^2\mathbb{I}^d$ covariance matrix are all coordinate-wise independent. 
To serve the distinguishing purpose: $\mathbf{z}$ was/was not used, it suffices to characterize the private gradient $p_i$ by $\bar{p}_i$ as a univariate random variable (the first coordinate) for the distinguishing purpose as follows.
\equienvs
\begin{equation}\label{equ:r1_e}
\begin{aligned}
    & \bar{p}_i|X\sim \mathcal{N}(0, C^2\sigma^2),\text{\quad}\bar{p}_i|X'\sim \mathcal{N}(C, C^2\sigma^2).
\end{aligned}
\end{equation}
\equienve

\textit{\ding{183} Second reduction} (from Equation \eqref{equ:r1_e} to Equation \eqref{equ:uni_game_tau_1}). Base on previous reduction, the DP-SGD's output $\bar{y} = \{\bar{p}_1,\bar{p}_2,\cdots,\bar{p}_N\}$ is essentially an observation of $N$ i.i.d. samples from  $\mathcal{N}(a, C^2\sigma^2)$ where $a$ is either $0$ or $C$. Recall the adversary's goal is to distinguish $X$ or $X'$ was used; this is equivalent to determining $a=0$ or $a=C$.

As both distributions in Equation \eqref{equ:r1_e} are Gaussian, 
we can use the \textit{sufficient statistics} for estimating $a$ \cite{fisher1922mathematical,DBLP:books/daglib/0016881}, which is the mean: $\bar{y} = \frac{1}{N}\sum_{i=1}^N \bar{p}_i$. Sufficient statistics do not lose any information for estimating $a$.
Finally, we can reduce the privacy of the base algorithm to an equivalent game for the adversary as 
\equienvs
\begin{equation}\label{equ:r2}
\begin{aligned}
    \bar{y}|X \sim \mathcal{N}(0, C^2\sigma^2/N) , \text{\quad} \bar{y}|X'\sim \mathcal{N}(C, C^2\sigma^2/N).
\end{aligned}
\end{equation}
\equienve
For simplicity, applying a simple invertible/lossless re-scaling gives us equivalent characterization: 
\equienvs
\begin{equation}\label{equ:uni_game_tau_1}
\begin{aligned}
    \bar{y}|X \sim \mathcal{N}(0, 1) , \text{\quad} \bar{y}|X'\sim \mathcal{N}(\sqrt{N}/\sigma, 1).
\end{aligned}
\end{equation}
\equienve

There is a slight difference in the reduction when $\tau<1$. Instead of arriving at Equation \eqref{equ:r1_e}, we arrive at
\equienvs
\begin{equation}\label{equ:r3_s}
\begin{aligned}
    \bar{p}_i|X \sim \mathcal{N}(0, C^2\sigma^2),\text{\quad} \bar{p}_i|X'\sim \mathcal{N}(Cb_i, C^2\sigma^2),
\end{aligned}
\end{equation}
\equienve
where $b_i, \forall i\in\{1,2,\cdots, N\}$ is independent Bernoulli random variables with probability $\tau$. By doing the same transformation as from Equation \eqref{equ:r1_e} to Equation \eqref{equ:uni_game_tau_1}, we arrive at 
\equienvs
\begin{equation}\label{equ:r2_accurate}
\begin{aligned}
    \bar{y}|X \sim \mathcal{N}(0, 1), \text{\quad} \bar{y}|X'\sim \mathcal{N}(\frac{\sum_{i=1}^N b_i}{N}\sqrt{N}/\sigma, 1).
\end{aligned}
\end{equation}
\equienve
Equation \eqref{equ:r2_accurate} also covers Equation \eqref{equ:uni_game_tau_1} when $\tau=1$.

\notbf{2) Instantiate the score function.} Based on our above reduction, to serve the distinguishing purpose, a model obtained by DP-SGD can be ``treated'' as a real number sampled from univariate Gaussian or its shifted counterpart corresponding to $X$ or $X'$ was used. The order induced by the score function $g$ is now over $\mathbb{R}$. To have stronger audit results in our idealized attack, we need to instantiate the worst-case score function, and our Theorem \ref{thm:bijective_large} tells us one-to-one mapping score function $g$ is the worst case necessarily. 

\begin{figure}[!t] 
    \centering
    {
    \includegraphics[width=.7\linewidth]{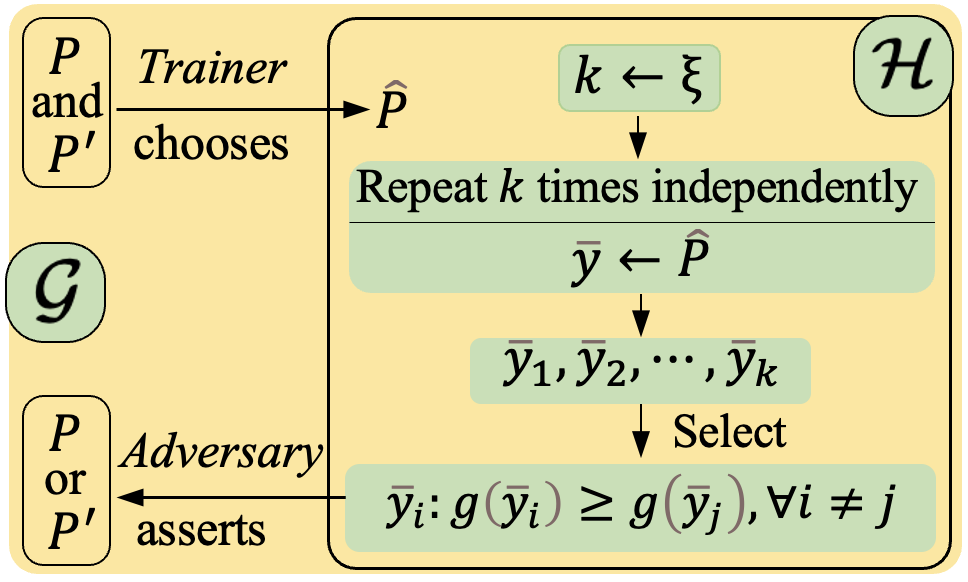}
    }
    \vspace{-0.1cm}
    \caption{
    The distinguishing game for $\mathcal{H}$ by reduction. $P,P'$ are two univariate Gaussians shown in Equation \eqref{equ:r2_accurate}, corresponding to adjacent dataset pair $X, X'$.
    The adversary then makes binary assertions by comparing the best sample to some threshold. The game $\mathcal{G}$ is simulated $10^7$ times.
    }
    \label{fig:equi_dis_game} 
    \vspace{\fmgctom}
\end{figure}

However, Theorem \ref{thm:bijective_large} remains silent on the specific analytical form of $g$ in the worst case. 
There can be infinitely many one-to-one mapping functions $\mathbb{R}\rightarrow\mathbb{R}$; for implementation purposes, we now fix a score function $g(x)=x$, i.e., we take $g$ is strictly increasing and note that all strictly increasing functions induces the same order over $\mathbb{R}$ regardless of its analytical form.
Finally, we present the distinguishing game of our reduced case in Figure \ref{fig:equi_dis_game}, which will be simulated many times, allowing high-confident conclusions of the lower bound following Section \ref{sec:audit_intro}.

\subsection{Privacy Results Comparison}\label{sec:bound_comparison}
\notbf{Privacy results.} We present the lower bound \textcolor{red}{$\varepsilon_L$}, \underline{\bf o}ur improved privacy results $\varepsilon^O_\mathcal{H}$ and \underline{\bf p}revious privacy results \textcolor{cyan}{$\varepsilon_\mathcal{H}^P$} \cite{papernot2021hyperparameter} in Table \ref{tab:result_comp}. We can see that the audit is rather powerful: \textcolor{red}{$\varepsilon_L$} $>\varepsilon_B$ with a noticeable margin, once again confirming the very action of selecting the best does leak additional privacy beyond the base algorithm's privacy budget. On the other hand, our privacy result $\varepsilon_\mathcal{H}^O$ is better than the previous result \textcolor{cyan}{$\varepsilon_\mathcal{H}^P$} under all setups.

\notbf{Implication.} The improvement we obtain is much more significant than the $\epsilon^O_\mathcal{H}$ value shows cosmetically. For instance, the results due to our improved upper bound $\epsilon^O_\mathcal{H}$ under $(\eta=1,\nu=10^{-3})$ are even consistently smaller than \textcolor{cyan}{$\varepsilon_\mathcal{H}^P$} of previous upper bound under $(\eta=1,\nu=10^{-2})$. Thus,
our improved analysis can allow trying significantly more hyper-parameter candidates while even consuming less privacy budget. For private hyper-parameter tuning applications, this translates to improved utility of the trained model for free.

\RevisionChangeStart
To show the downstream gain using our method, we present the privacy result comparison in Table \ref{tab:privacy_compare}. As can be seen from Table \ref{tab:privacy_compare}, under the same privacy constraint, we allow more running (in expectation) hyper-parameter candidates, and the final achievable testing accuracy is consistently better as expected.
\RevisionChangeEnd

\begin{table}[!t] 
\Large
\centering

\resizebox{1.0\columnwidth}{0.5\height}{
\begin{tabular}{cc|c|c|c|c}

\toprule

$\varepsilon_B$
&
$\tau$
& 
\multicolumn{4}{c}{
\textcolor{red}{$\varepsilon_L$} | $\varepsilon_\mathcal{H}^O$ | \textcolor{cyan}{$\varepsilon_\mathcal{H}^P$}
}
\\

\midrule
&
& 
\makecell[c]{
$\eta=0, \nu=10^{-2}$\\
$\mathbb{E}_\xi=21$
} 
& 
\makecell[c]{
$\eta=1, \nu=10^{-2}$\\
$\mathbb{E}_\xi=100$
}
& 
\makecell[c]{
$\eta=1, \nu=10^{-3}$\\
$\mathbb{E}_\xi=1000$
}
& 
\makecell[c]{
$\eta=2, \nu=10^{-3}$\\
$\mathbb{E}_\xi=2000$
}
\\
&&&&\\[-5pt]

\multirow[t]{3}{*}[-13pt]{1}
&
1
&
\makecell[l]{
\textcolor{red}{$1.17$} | $1.55$ | \textcolor{cyan}{$1.86$}
}
&
\makecell[l]{
\textcolor{red}{$1.19$} | $2.06$ | \textcolor{cyan}{$2.65$}
}
&
\makecell[l]{
\textcolor{red}{$1.27$} | $2.54$ | \textcolor{cyan}{$3.09$}
}
&
\makecell[l]{
\textcolor{red}{$1.29$} | $3.18$ | \textcolor{cyan}{$3.99$}
}
\\
&
0.5
&
\makecell[l]{
\textcolor{red}{$1.09$} | $1.53$ | \textcolor{cyan}{$1.87$}
}
&
\makecell[l]{
\textcolor{red}{$1.30$} | $2.04$ | \textcolor{cyan}{$2.64$}
}
&
\makecell[l]{
\textcolor{red}{$1.37$} | $2.51$ | \textcolor{cyan}{$3.15$}
}
&
\makecell[l]{
\textcolor{red}{$1.42$} | $3.15$ | \textcolor{cyan}{$4.05$}
}
\\
&
0.1
&
\makecell[l]{
\textcolor{red}{$1.00$} | $1.49$ | \textcolor{cyan}{$1.86$}
}
&
\makecell[l]{
\textcolor{red}{$1.19$} | $1.98$ | \textcolor{cyan}{$2.56$}
}
&
\makecell[l]{
\textcolor{red}{$1.49$} | $2.43$ | \textcolor{cyan}{$3.24$}
}
&
\makecell[l]{
\textcolor{red}{$1.46$} | $3.05$ | \textcolor{cyan}{$4.09$}
}
\\
\midrule
\multirow[t]{3}{*}[-13pt]{2}
&
1
&
\makecell[l]{
\textcolor{red}{$2.17$} | $2.92$ | \textcolor{cyan}{$3.61$}
}
&
\makecell[l]{
\textcolor{red}{$2.21$} | $3.84$ | \textcolor{cyan}{$5.06$}
}
&
\makecell[l]{
\textcolor{red}{$2.53$} | $4.69$ | \textcolor{cyan}{$5.89$}
}
&
\makecell[l]{
\textcolor{red}{$2.57$} | $5.85$ | \textcolor{cyan}{$7.57$}
}
\\
&
0.5
&
\makecell[l]{
\textcolor{red}{$2.05$} | $2.89$ | \textcolor{cyan}{$3.57$}
}
&
\makecell[l]{
\textcolor{red}{$2.34$} | $3.80$ | \textcolor{cyan}{$4.93$}
}
&
\makecell[l]{
\textcolor{red}{$2.59$} | $4.64$ | \textcolor{cyan}{$5.88$}
}
&
\makecell[l]{
\textcolor{red}{$2.63$} | $5.79$ | \textcolor{cyan}{$7.49$}
}
\\
&
0.1
&
\makecell[l]{
\textcolor{red}{$2.16$} | $2.90$ | \textcolor{cyan}{$3.58$}
}
&
\makecell[l]{
\textcolor{red}{$2.40$} | $3.82$ | \textcolor{cyan}{$4.83$}
}
&
\makecell[l]{
\textcolor{red}{$2.87$} | $4.67$ | \textcolor{cyan}{$6.03$}
}
&
\makecell[l]{
\textcolor{red}{$2.86$} | $5.82$ | \textcolor{cyan}{$7.53$}
}
\\
\midrule
\multirow[t]{3}{*}[-13pt]{4}
&
1
&
\makecell[l]{
\textcolor{red}{$3.93$} | $5.70$ | \textcolor{cyan}{$6.80$}
}
&
\makecell[l]{
\textcolor{red}{$4.32$} | $7.40$ | \textcolor{cyan}{$9.30$}
}
&
\makecell[l]{
\textcolor{red}{$4.51$} | $8.95$ | \textcolor{cyan}{$10.83$}
}
&
\makecell[l]{
\textcolor{red}{$4.70$} | $11.07$ | \textcolor{cyan}{$13.77$}
}
\\
&
0.5
&
\makecell[l]{
\textcolor{red}{$3.84$} | $5.64$ | \textcolor{cyan}{$6.71$}
}
&
\makecell[l]{
\textcolor{red}{$4.23$} | $7.32$ | \textcolor{cyan}{$9.03$}
}
&
\makecell[l]{
\textcolor{red}{$4.89$} | $8.86$ | \textcolor{cyan}{$10.74$}
}
&
\makecell[l]{
\textcolor{red}{$4.91$} | $10.96$ | \textcolor{cyan}{$13.51$}
}
\\
&
0.1
&
\makecell[l]{
\textcolor{red}{$3.76$} | $5.48$ | \textcolor{cyan}{$6.58$}
}
&
\makecell[l]{
\textcolor{red}{$4.17$} | $7.12$ | \textcolor{cyan}{$8.64$}
}
&
\makecell[l]{
\textcolor{red}{$5.02$} | $8.62$ | \textcolor{cyan}{$10.61$}
}
&
\makecell[l]{
\textcolor{red}{$5.03$} | $10.67$ | \textcolor{cyan}{$13.08$}
}
\\

\bottomrule
\end{tabular}

}
\caption{
\RevisionChangeStart
Comparison of privacy bounds for $\mathcal{H}$ (Algorithm~\ref{alg:ps}). All values are in $(\varepsilon, \delta = 10^{-5})$-DP form. 
\textcolor{red}{$\varepsilon_L$} is the empirical lower bound obtained by simulating the distinguishing game in Figure~\ref{fig:equi_dis_game}. 
$\varepsilon_\mathcal{H}^O$ is our improved analytical upper bound. 
\textcolor{cyan}{$\varepsilon_\mathcal{H}^P$} is the upper bound from prior work~\cite{papernot2021hyperparameter}.Each row corresponds to a different sampling ratio $\tau$, with total iterations fixed at $N = 10^3$. 
The parameters $\eta$ and $\nu$ define the TNB distribution used to generate the number of runs in $\mathcal{H}$ (details in Appendix~\ref{app:TNBD}), and $\mathbb{E}_\xi$ is the expected number of runs under this distribution.
\RevisionChangeEnd
}

\label{tab:result_comp}
\end{table}

\begin{table}[!t] 
\Large
\centering

\resizebox{1.0\columnwidth}{0.55\height}{
\begin{tabular}{cc|c|c|c|c}

\toprule

$\varepsilon_B$
& 
$\varepsilon^O_\mathcal{H}$
&
\multicolumn{4}{c}{
    Previous $\rightarrow$ Ours
}
\\

\midrule

&
& 
MNIST
& 
FMNIST
& 
CIFAR10
& 
SVHN
\\
&&&&
\\[-8pt]

$1$
& 
$1.83$
& 
$0.921$  $\rightarrow$ $0.934$
& 
$0.768$  $\rightarrow$ $0.793$
& 
$0.412$ $\rightarrow$ $0.448$
& 
$0.636$  $\rightarrow$ $0.661$
\\

$2$
& 
$3.43$
& 
$0.942$  $\rightarrow$ $0.956$
& 
$0.779$  $\rightarrow$ $0.802$
& 
$0.467$ $\rightarrow$ $0.486$
& 
$0.706$  $\rightarrow$ $0.745$
\\

$4$
& 
$6.69$
&
$0.951$  $\rightarrow$ $0.958$
& 
$0.791$  $\rightarrow$ $0.817$
& 
$0.504$ $\rightarrow$ $0.531$
& 
$0.762$  $\rightarrow$ $0.786$
\\

\bottomrule
\end{tabular}

}
\caption{
\RevisionChangeStart
Testing accuracy comparison under differentially private hyper-parameter tuning. The TNB setup for our method is $(\eta=0,\nu=10^{-3})$ ($\mathbb{E}_\xi\approx 144$). To achieve roughly the same privacy result using previous method \cite{papernot2021hyperparameter}, the setup should be $(\eta=0,\nu=10^{-2})$ ($\mathbb{E}_\xi\approx 21$). $\varepsilon^O_\mathcal{H}$ is our improved result for private hyper-parameter tuning. For all experiments, $\delta=10^{-5}$.
\RevisionChangeEnd
}

\label{tab:privacy_compare}
\vspace{-.5cm}
\end{table}

\subsection{Lessons Learned and Open Problems}
As shown in Section \ref{sec:bound_disc}, our improved result is indeed tight for general $\xi$ in terms of how much $(\varepsilon_\mathcal{H},\delta_\mathcal{H})$-DP $\mathcal{H}$ satisfies. However, we still see a noticeable gap between our result $\varepsilon^O_\mathcal{H}$ and the lower bound \textcolor{red}{$\varepsilon_L$}  derived by our idealized attack. why does this happen? 

Our answer is that $g$ plays a critical role from the adversary's point of view, and such a factor distinguishes attacking private hyper-parameter tuning from all previous privacy attack problems. We have shown that one-to-one mapping is necessary for $g$ being the worst case, but there are infinitely many $g$ over an infinite output domain and are up to $|\Gamma|!$ possible choices if output domain $\Gamma$ is finite. We find that some of the score functions leak more privacy than others. For example, when $\eta=1, \nu=10^{-2}$ (TNB $\xi$ recovers geometric distribution), if we (arbitrarily) set the score function as 
\equienvs
\begin{equation}
    g(x)=\left\{\begin{array}{ll}
        x, \hspace{0.56cm}\text{for $x\in[-\infty,0) \bigcup (1,\infty]$}\\
        1 - x, \text{for $x\in[0,1]$}\\
    \end{array}\right.
\end{equation}
\equienve
which is clearly a one-to-one mapping, we only derive \textcolor{red}{$\varepsilon_L=2.01$} (average of 10 runs) at $\varepsilon_B=2, \tau=1$, which is smaller/weaker than the value $2.21$ shown in Table \ref{tab:result_comp} (where $g(x)=x$).

Choosing some $g$ arbitrarily and performing the attack will likely end up with sub-optimal attacks (smaller/weaker lower bounds). This is probably the reason why we still see a gap between $\varepsilon^O_\mathcal{H}$ and \textcolor{red}{$\varepsilon_L$} even in our idealized attack. Reasoning on such issues is non-trivial, and it poses the following questions worthy of investigation:

1) which $g$ should the adversary choose to elicit more privacy leakage? 2) does the worst-case $g$ depend on specific $\xi$? 3) How to quantify the exact trade-off between privacy leakage and $\xi$ which governs the utility? Answering the above questions requires non-trivial efforts, which we hold as meaningful future directions.

\section{Conclusion}

We study how selection leaks privacy. Initially, we give examples showing that the current generic bound for private selection is indeed tight in general. Still, it is not tight for a white-box setting, i.e., the hyper-parameter tuning problem. Substantiating this assertion, we first audit the privacy of hyper-parameter tuning under various settings; the derived empirical privacy lower bound under the strongest adversary still sees a noticeable gap from the generic upper bound.

We then provide an in-depth study of deriving better privacy upper bound by modeling the base algorithm's privacy with finer resolution ($f$-DP). 
The improvement is due to the distinct properties of the base algorithm (DP-SGD). 
Our result allows trying many more hyper-parameter candidates while consuming less private budget. Our analysis also generalizes, contrasting with previous work, which remains unknown how to adapt to general parameter setups.

\section{Ethics Considerations}
This paper is on refining the privacy bound for differentially private protocols, not on privacy attacks. The privacy audit experiments conducted are to conclude a privacy lower bound, not to launch some real-world privacy attacks. All analyses and experiments are conducted using publicly available datasets to minimize privacy risks. Our study aims to strengthen differential privacy protections in hyper-parameter tuning by improving the analysis rather than exploiting any weaknesses.

\bibliographystyle{plain}
\bibliography{ref}

\appendix

\section{Content for reference}
\subsection{Neyman–Pearson Lemma}\label{app:np_lemma}
\begin{theorem}[Neyman–Pearson lemma \cite{neyman1933ix}]
Let $P$ and $Q$ be probability distributions on $\Omega$ with densities $p$ and $q$, respectively. Define $L(x)=\frac{p(x)}{q(x)}$. For hypothesis testing problem 
{\small$$\mathbf{H_0}:P,\text{\quad}\mathbf{H_1}: Q$$}For a constant $c>0$, suppose that the likelihood ratio test which rejects $\mathbf{H_0}$ when $L(x)\leq c$ has $\mathrm{FP}=a$ and $\mathrm{FN}=b$, then for any other test of $\mathbf{H_0}$ with $\mathrm{FP}\leq a$, the achievable $\mathrm{FN}$ is at least $b$.
    
\end{theorem}

Neyman–Pearson lemma says that the most powerful test (optimal $\mathrm{FN}$) at fixed  $\mathrm{FP}$ is the likelihood ratio test. Applying Neyman–Pearson lemma to distinguishing $\mathcal{N}(0,1)$ from $\mathcal{N}(\mu,1)$ gives us Definition \ref{def:gdp} \cite{dong2019gaussian}.

\subsection{Privacy Results by Papernot et al.}\label{app:TNBD}
\notbf{Truncated negative binomial (TNB) distribution.} For $\nu\in(0,1)$ and $\eta\in(-1,\infty)$, the distribution $\xi_{\eta,\nu}$ on $\{1,2,3,\cdots\}$ is as follows. When $\eta\neq0$, then 
\equienvs
\begin{equation}\nonumber
\begin{aligned}
    \forall k \in \mathbb{N}, \quad \operatorname{Pr}[K=k]=\frac{(1-\nu)^k}{\nu^{-\eta}-1} \cdot \prod_{\ell=0}^{k-1}\left(\frac{\ell+\eta}{\ell+1}\right),
\end{aligned}
\end{equation}
\equienve
when $\eta=0$, then 
\equienvs
\begin{equation}\nonumber
\begin{aligned}
    \forall k \in \mathbb{N}, \quad \operatorname{Pr}[K=k]=\frac{(1-\nu)^k}{k \cdot \log (1 / \nu)}
\end{aligned}
\end{equation}
\equienve
This particular distribution is obtained by differentiating the probability generating function of some desired form \cite{papernot2021hyperparameter}. 
The main relevant privacy results in \cite{papernot2021hyperparameter} are provided in the following. Note that they are all in RDP form.

\begin{theorem}[RDP for TNB distribution \cite{papernot2021hyperparameter}]
    Let $k$ in Algorithm \ref{alg:audit} follows TNB distribution $\xi_{\eta,\nu}$. If the base algorithm satisfies $(\alpha,\gamma)$-RDP and  $(\alpha',\gamma')$-RDP, Algorithm \ref{alg:audit} satisfies $(\alpha, \hat{\gamma})$-RDP where
    \equienvs
    \begin{equation}\nonumber
    \begin{aligned}
        \hat{\gamma}=\gamma +(1+\eta) \cdot\left(1-\frac{1}{\alpha'}\right) \gamma'+\frac{(1+\eta) \cdot \log (1 / \nu)}{\alpha'}+\frac{\log \mathbb{E}_{\xi_{\eta,\nu}}}{\alpha-1}
    \end{aligned}
    \end{equation}
    \equienve
\end{theorem}

\notbf{Tight example for approximate DP.}
A Tight example for $(\varepsilon,\delta)$-DP case can be obtained trivially based on Example \ref{example:pure_dp_tight}. If an algorithm is $(\varepsilon,0)$-DP, it is also $(\varepsilon,\delta)$-DP. Hence, the above tight example covers the $(\varepsilon,\delta)$-DP case. Specifically, it can be checked that the example shown in Equation \eqref{equ:base_pure_dp} is $(1,10^{-5})$-DP and Equation \eqref{equ:hyper_pure_dp} is $(2.92,10^{-5})$-DP. Compared to the result predicted by \cite{papernot2021hyperparameter}, which is $(3.11,10^{-5})$-DP, i.e., it is tight up to a negligible gap.

\subsection{Used Datasets and Experimental Details}\label{app:exp_detail}
Our implementation is provided at an anonymous link\footnote{{https://github.com/zihangxiang/PrivateHyperparameterTuning.git}}. We use four image datasets in our experiments. 
FASHION \cite{xiao2017/online}, MNIST \cite{lecun1998gradient}, CIFAR10 \cite{krizhevsky2009learning} and SVHN \cite{netzer2011reading}.
All of our experiments are conducted under privacy parameter $\delta=10^{-5}$. 
The number of repeating/simulation times in an audit experiment is $2,000$. The error bar results from taking the min., max., and avg. for three trials. To efficiently audit the hyper-parameter tuning and reduce the simulation burden, we only fetch 5,000 data examples from the original training datasets, and we set the sampling rate to be 1, i.e., full gradient descent. We set the TNB distribution \cite{papernot2021hyperparameter} with parameter $(\eta=0,\nu=10^{-2})$.
We use the ResNet \cite{HeZRS16} as the neural network in our experiments. We use \textit{Adam} as the default optimizer.
The computational burden is significant: our audit experiment consumes $>4000$ GPU hours and is conducted over 20 GPUs in parallel. 

\notbf{Hyperparameter candidates setup.} To run the audit experiments, we need to set the candidates inside $\Omega$ in Algorithm \ref{alg:ps}. We hold the clipping threshold $C$, learning rate $\mathbf{lr}$, and the number of total iterations $N$ as the hyperparameters to be tuned. To form each candidate inside $\Omega$, we randomly sample a value to determine $C$, $\mathbf{lr}$ and $N$. 
All candidates have the same privacy budget according to our problem formulation.  

\notbf{Detailed procedure of concluding the lower bound $\varepsilon_L$}. The following procedure is adopted to conclude a lower bound $\varepsilon_L$.

1) \textbf{Generating $(b_{\text{truth}},b_{\text{guess}})$}. Each pair corresponds to an execution of $\mathcal{G}$ (Algorithm \ref{alg:audit}). The adversary needs to make an assertion, 
i.e., output a $b_{\text{guess}}\in\{0,1\}$. 

2) \textbf{Compute $\varepsilon_L$}. After getting many pairs of $(b_{\text{truth}},b_{\text{guess}})$, the $\mathrm{FP}, \mathrm{FN}$  can be summarised by Clopper-Pearson with a confidence $c$. Specifically, the $\mathrm{FP}$ rate and $\mathrm{FN}$ rate are modeled as the unknown success probabilities of two binomial distributions. Then $\varepsilon_L$ can be computed by Equation \eqref{equ:lower_bound} or by the methods used in \cite{nasr2023tight}.

3) \textbf{Optimization}. In practice, practitioners often try various assertion strategies on the same observed output by repeating procedures 1) and 2) to find the optimal $\varepsilon_L$.

\section{Proofs}

\subsection{Proof of Claim \ref{thm:prob_compute}}\label{app:proof_of_prob_compute}
\begin{proof}
    when $k>0$ is some fixed integer, we know that the scores of all $k$ runs are $\leq g(Y)$, which has the probability $\operatorname{Pr}(E_{\leq y})^k$. As $y$ occurs, we have probability $\operatorname{Pr}(E_{\leq y})^k-\operatorname{Pr}(E_{< y})^k$ seeing $y$ as the output of Algorithm \ref{alg:ps}. When $k$ follows some general distribution $\xi$, the resultant distribution is a mixture, which is Claim \ref{thm:prob_compute}.
\end{proof}

\subsection{Proof of Theorem \ref{thm:bijective_large}}\label{app:proof_bijective_large}

\begin{proof}

W.o.l.g., we define the alphabets of distribution $P$ and $P'$ as $\{a,b,c,d,e,f\}$, with some abuse of notation, we denote
\equienvs
\begin{equation}\nonumber
\begin{aligned}
    &P(a) = p_a, P(b) = p_b,\cdots, P(f)=p_f\\
    &P'(a) = p'_a, P'(b) = p'_b,\cdots, P'(f)=p'_f\\
\end{aligned}
\end{equation}
\equienve
as their probabilities. Suppose we have a non-one-to-one mapping score evaluator $\hat{g}$ such that:
\equienvs
\begin{equation}\nonumber
\begin{aligned}
    &\hat{g}(a) = \hat{g}(c) = \hat{g}(e)< \hat{g}(b)= \hat{g}(d) < \hat{g}(f).\\
\end{aligned}
\end{equation}
\equienve
We now assume a uniformly random selection among Alphabets that share the same score. 
For clearer presentation, we denote $\Lambda_S^k=(\sum_{i\in S}p_i)^k$ and $\Bar{\Lambda}_S^k=(\sum_{i\in S}p'_i)^k$. Then, the distribution of $F_{k, \hat{g}}$ will be
\equienvs
\begin{equation}\nonumber
\begin{aligned}
    &F_{k, \hat{g}}(a) = F_{k, \hat{g}}(c) = F_{k, \hat{g}}(e)=\frac{1}{3}\Lambda_{\{a,c,e\}}^k\\
    &F_{k, \hat{g}}(b) = F_{k, \hat{g}}(d) = \frac{1}{2}(\Lambda_{\{a,c,e,b,d\}}^k-\Lambda_{\{a,c,e\}}^k)\\
    &F_{k, \hat{g}}(f) = 1 - \Lambda_{\{a,c,e,b,d\}}^k
\end{aligned}
\end{equation}
\equienve
This is because
\equienvs
\begin{equation}\nonumber
\begin{aligned}
    &F_{k, \hat{g}}(i\in\{a,c,e\})=\Lambda_{\{a,c,e\}}^k\\
    &F_{k, \hat{g}}(i\in\{b,d\})=\Lambda_{\{a,c,e,b,d\}}^k-\Lambda_{\{a,c,e\}}^k
\end{aligned}
\end{equation}
\equienve
and a uniformly random selection among $\{a,c,e\}$ means that the probability mass $F_{k, \hat{g}}(i\in\{a,c,e\})$ is distributed uniformly to each. Similarly, the distribution of $F'_{k, \hat{g}}$ corresponding to $P'$ has the same form (just replace $\Lambda_S^k$ by $\Bar{\Lambda}_S^k$). We now construct a one-to-one mapping score function $g^*$ as follows.
\equienvs
\begin{equation}\nonumber
\begin{aligned}
    &g^*(a) < g^*(c) < g^*(e)<g^*(b) < g^*(d)< g^*(f)\\
\end{aligned}
\end{equation}
\equienve
The key point here is to \textbf{enforce a strict total order for alphabets that have the same score}. Then, the distribution of $F_{k, g^*}$ is
\equienvs
\begin{equation}\nonumber
\begin{aligned}
    &F_{k, g^*}(a) = \Lambda_{\{a\}}^k, F_{k, g^*}(c) = \Lambda_{\{a,c\}}^k- \Lambda_{\{a\}}^k\\
    &F_{k, g^*}(e) = \Lambda_{\{a,c,e\}}^k- \Lambda_{\{a,c\}}^k, F_{k, g^*}(b) = \Lambda_{\{a,c,e,b\}}^k- \Lambda_{\{a,c,e\}}^k\\
    &F_{k, g^*}(d) = \Lambda_{\{a,c,e,b,d\}}^k- \Lambda_{\{a,c,e,b\}}^k, F_{k, g^*}(f) = 1- \Lambda_{\{a,c,e,b,d\}}^k\\
\end{aligned}
\end{equation}
\equienve
Similarly, the distribution of $F'_{k, g^*}$ corresponding to $P'$ has the same form (just replace $\Lambda_S^k$ by $\Bar{\Lambda}_S^k$). We now compute the RDP quantity $\mathcal{D}_{\alpha}(F_{k, \hat{g}}||F'_{k,\hat{g}})$ and $\mathcal{D}_{\alpha}(F_{k, g^*}||F'_{k,g^*})$. We aim to show that the RDP value under non-one-to-one mapping $\hat{g}$ is smaller than that under its one-to-one mapping counterpart. We group the sub-terms of RDP calculation. Let 
\equienvs
\begin{equation}\nonumber
\begin{aligned}
    &\hat{T}_{\{a,c,e\}} = \sum_{i\in\{a,c,e\}}\left(\frac{F_{k, \hat{g}}(i)}{F'_{k, \hat{g}}(i)}\right)^\alpha F'_{k, \hat{g}}(i)\\
    &\hat{T}_{\{b,d\}} = \sum_{i\in\{b,d\}}\left(\frac{F_{k, \hat{g}}(i)}{F'_{k, \hat{g}}(i)}\right)^\alpha F'_{k, \hat{g}}(i)\\
    &\hat{T}_{\{f\}} = \left(\frac{F_{k, \hat{g}}(f)}{F'_{k, \hat{g}}(f)}\right)^\alpha F'_{k, \hat{g}}(f)
\end{aligned}
\end{equation}
\equienve
We compute the $T^*_{\{a,c,e\}}, T^*_{\{b,d\}}, T^*_{\{f\}}$ counterparts in the same fashion (just replace $\hat{g}$ by $g^*$). And we will compare $T_{\{a,c,e\}}$ and $T'_{\{a,c,e\}}$. 
By letting
\equienvs
\begin{equation}\nonumber
\begin{aligned}
    x = \Lambda_{\{a\}}^k \text{ \quad\quad\quad\quad\quad\quad\quad} &x' = \Bar{\Lambda}_{\{a\}}^k\\
    y = \Lambda_{\{a,c\}}^k- \Lambda_{\{a\}}^k \text{ \quad\quad\quad } &y'=\Bar{\Lambda}_{\{a,c\}}^k - \Bar{\Lambda}_{\{a\}}^k\\
    z= \Lambda_{\{a,c,e\}}^k -\Lambda_{\{a,c\}}^k \text{ \quad\quad } &z' = \Bar{\Lambda}_{\{a,c,e\}}^k -\Bar{\Lambda}_{\{a,c\}}^k
\end{aligned}
\end{equation}
\equienve
then, it is easy to see that 
\equienvs
\begin{equation}\label{equ:critical_sum}
\begin{aligned}
    \hat{T}_{\{a,c,e\}} = &(\frac{\Lambda_{\{a,c,e\}}^k}{\Bar{\Lambda}_{\{a,c,e\}}^k})^\alpha \Bar{\Lambda}_{\{a,c,e\}}^k\\
    = & (\frac{x+y+z}{x'+y'+z'})^\alpha(x'+y'+z')\\
    \leq & (\frac{x}{x'})^\alpha x'+(\frac{y}{y'})^\alpha y'+(\frac{z}{z'})^\alpha z'\\
    = & T^*_{a,c,e} 
\end{aligned}
\end{equation}
\equienve
holds by Jensen's inequality and the fact that function $h(x)=x^\alpha$ is convex for $\alpha>1$, i.e.,
\equienvs
\begin{equation}\nonumber
\begin{aligned}
    (\frac{x+y+z}{x'+y'+z'})^\alpha\leq  \frac{(\frac{x}{x'})^\alpha x'+(\frac{y}{y'})^\alpha y'+(\frac{z}{z'})^\alpha z'}{x'+y'+z'}\\
\end{aligned}
\end{equation}
\equienve
For the same reason, it can also be easily checked that $\hat{T}_{\{b,d\}}\leq T^*_{\{b,d\}}$ and $\hat{T}_{\{f\}}\leq T^*_{\{f\}}$ also hold. Because
\equienvs
\begin{equation}\nonumber
\begin{aligned}
    &\mathcal{D}_{\alpha}(F_{k, \hat{g}}||F'_{k,\hat{g}}) = \frac{1}{\alpha-1}\ln(\hat{T}_{\{a,c,e\}} + \hat{T}_{\{b,d\}}+ \hat{T}_{\{f\}})\\
    &\mathcal{D}_{\alpha}(F_{k, g^*}||F'_{k,g^*}) =\frac{1}{\alpha-1}\ln(T^*_{\{a,c,e\}} + T^*_{\{b,d\}}+ T^*_{\{f\}}),
\end{aligned}
\end{equation}
\equienve
we have $$\mathcal{D}_{\alpha}(F_{k, \hat{g}}||F'_{k,\hat{g}}) \leq \mathcal{D}_{\alpha}(F_{k, g^*}||F'_{k,g^*}).$$ Note the first equality of Equation \eqref{equ:critical_sum} always holds no matter whether selection among alphabets sharing the same score is uniform or weighted; the alphabet and the order we choose is also arbitrary, which means that the result holds in general.

\notbf{Remark.} Following the same reasoning, when $k$ is now a random variable instead of a fixed number, we also have the result, as shown in the above theorem. Because we can modify each probability term to be the probability of the mixture counterpart, and the proof follows trivially. Specifically, for each probability $p=f(k)$ shows up, modify it to be $p=\sum_{i=1}^{\infty} \operatorname{Pr}(i)f(i)$ where $\operatorname{Pr}(i), i=\{1,2,\cdots,\infty\}$ is the p.m.f. of distribution $\xi$.
\end{proof}

\subsection{Proof of Theorem \ref{thm:improved_privacy_upper_bound}}\label{app:proof_of_improved_privacy_upper_bound}
\begin{proof}
    As we care about how much $(\varepsilon_\mathcal{H},\delta_\mathcal{H})$-DP $\mathcal{H}$ satisfies given some $\delta_\mathcal{H}$, it is useful to introduce a technical lemma related to such form of DP.
    \begin{lemma}[\cite{steinke2022composition} Propostion 7]\label{lem:equ_eps_delta}
        Define the privacy loss random variable for a pair of adjacent dataset $X,X'$ to a private mechanism $\mathcal{M}$ as $L_1=\log\frac{\mathcal{M}(X)(o)}{\mathcal{M}(X')(o)}$ where $o\sim\mathcal{M}(X)$ $\mathcal{M}$ is $(\varepsilon,\delta)$-DP or $f_{\varepsilon,\delta}$-DP if and only if
        $$\int_\varepsilon^\infty\mathrm{e}^{\varepsilon-z}\cdot \operatorname{Pr}_{o\sim\mathcal{M}(X)}[L_1>z]\mathrm{d}z\leq \delta $$
        holds for all adjacent $X,X'$.    
    \end{lemma}
    Our goal is clear, i.e., we need to meet the following equation:
    \equienvs
    \begin{equation}\label{equ:goal}
    \begin{aligned}
        \int_{\varepsilon_\mathcal{H}}^\infty\mathrm{e}^{\varepsilon_\mathcal{H}-z}\cdot \operatorname{Pr}_{o\sim Q}[\log\frac{Q(o)}{Q'(o)}>z]\mathrm{d}z\leq \delta_\mathcal{H}
    \end{aligned}
    \end{equation}
    \equienve
    Then $\mathcal{H}$ would be $(\varepsilon_\mathcal{H},\delta_\mathcal{H})$-DP.

    Let the left-hand side of Equation \eqref{equ:goal} to be $t_{\varepsilon_\mathcal{H}}$ and note that $\frac{Q(o)}{Q'(o)}=\frac{P(o)\omega_\xi(F(o))}{P'(o)\omega_\xi(F'(o))}$, define event $E_z=\{o:\log\frac{P(o)\omega_\xi(F(o))}{P'(o)\omega_\xi(F'(o))}>z\}$ then
    \equienvs
    \begin{equation}\label{equ_t_eps_h}
    \begin{aligned}
        t_{\varepsilon_\mathcal{H}} =&\int_{\varepsilon_\mathcal{H}}^\infty\mathrm{e}^{\varepsilon_\mathcal{H}-z}\cdot \int\limits_{E_z} Q(o)\mathrm{d}o\mathrm{d}z\\
        \leq& \omega_\xi(1)\int_{\varepsilon_\mathcal{H}}^\infty\mathrm{e}^{\varepsilon_\mathcal{H}-z}\cdot \int\limits_{E_z} P(o)\mathrm{d}o\mathrm{d}z
    \end{aligned}
    \end{equation}
    \equienve
    The inequality is due to $\omega_\xi:[0,1]\rightarrow \mathbb{R}$ is increasing. 
    
    Let us investigate the hypothesis testing problem $P$ V.S. $P'$, i.e., deciding $X$ or $X'$ was used based on the score of a single run of the DP-SGD. The score is post-processing~\cite[Lemma 1]{dong2019gaussian}) of the trained model, so the $(\mathrm{FP},\mathrm{FN})$ pair for distinguishing $P$ from $P'$ is governed by $f$.
    
    For some real number $o\in\mathbb{R}$, define $A=\{u:u\leq o\}$, and a decision rule $\mathcal{R}$ that accepts $P$ when the score falls into $A$. Then, $\mathrm{FP}_\mathcal{R}=1-F(o)$ and $\mathrm{FN}_\mathcal{R}=F'(o)$. And we must have $F'(o)\geq f(1-F(o))$ as governed by the trade-off function. This leads to an upper bound (note that $\omega_\xi$ is increasing)
    \equienvs
    \begin{equation}\label{equ:ratio_upper_bound}
    \begin{aligned}
        \frac{\omega_\xi(F(o))}{\omega_\xi(F'(o))}\leq \max_{a\in [0,1]}{\frac{\omega_\xi(1-a)}{\omega_\xi(f(a))}}=M
    \end{aligned}
    \end{equation}
    \equienve
    Now, let $\hat{E}_z=\{o:\log\frac{P(o)}{P'(o)}M>z\}$, it is easy to see that $E_z\subseteq\hat{E}_z$. 
    Hence, we have 
    \equienvs
    \begin{equation}\nonumber
    \begin{aligned}
        t_{\varepsilon_\mathcal{H}} \leq& \omega_\xi(1)\int_{\varepsilon_\mathcal{H}}^\infty\mathrm{e}^{\varepsilon_\mathcal{H}-z}\cdot \int\limits_{\hat{E}_z} P(o)\mathrm{d}o\mathrm{d}z\\
        =&\omega_\xi(1)\int_{\varepsilon_\mathcal{H}}^\infty\mathrm{e}^{\varepsilon_\mathcal{H}-z}\cdot \operatorname{Pr}_{o\sim P}[\log\frac{P(o)}{P'(o)}>z-\log M]\mathrm{d}z\\
        \leq&\omega_\xi(1)\int_{\varepsilon_\mathcal{H}-\log M}^\infty\mathrm{e}^{\varepsilon_\mathcal{H}-z}\cdot \operatorname{Pr}_{o\sim P}[\log\frac{P(o)}{P'(o)}>z-\log M]\mathrm{d}z\\
        =&\omega_\xi(1)\int_{\varepsilon_\mathcal{H}-\log M}^\infty\mathrm{e}^{\varepsilon_\mathcal{H}-\log M -s}\cdot \operatorname{Pr}_{o\sim P}[\log\frac{P(o)}{P'(o)}>s]\mathrm{d}s
    \end{aligned}
    \end{equation}
    \equienve
    Letting $s=z-\log M$, we have the last equality. Note that the score is differentially private, as it is post-processing of the base algorithm. Hence, we can compute a $(\varepsilon_\mathcal{H}-\log M, \delta)$-DP guarantee for the score. Applying Lemma \ref{lem:equ_eps_delta}, we have $$t_{\varepsilon_\mathcal{H}}\leq \omega_\xi(1)\delta.$$
    Setting $\delta_\mathcal{H}=\omega_\xi(1)\delta$, we derive $\delta$. By inputting trade-off function $f$ for the base algorithm and $\delta$ to Algorithm \ref{alg:fdp_to_eps_delta}, we derive the value of $\epsilon_\mathcal{H}-\log{M}$, which give us Theorem \ref{thm:improved_privacy_upper_bound}. As we assume nothing on the adjacent dataset $X,X'$, Theorem \ref{thm:improved_privacy_upper_bound} holds for all $X,X'$ pair.

\end{proof}






%



\end{document}